\documentclass[11pt, a4paper]{report} 

\usepackage[utf8]{inputenc}
\usepackage[margin=2cm]{geometry}
\usepackage{graphicx}
\usepackage{amsmath}
\usepackage{array}
\usepackage{enumerate}
\usepackage{amsthm}
\usepackage{bbold}

\renewcommand{\vec}[1]{\textbf{#1}}
\usepackage{amssymb}
\usepackage{hyperref}
\usepackage{parskip}
\usepackage{float}
\usepackage{booktabs}
\usepackage{caption}
\usepackage{lipsum}
\usepackage{subfig} 
\usepackage{cases}
\usepackage{algorithm2e}
\RestyleAlgo{ruled}

\usepackage{setspace}
\usepackage[
backend=bibtex,
style=numeric,
]{biblatex}
\usepackage{listings}
\usepackage{xcolor}
\lstdefinestyle{matlabcode}{
	backgroundcolor=\color{gray!10},
	commentstyle=\color{green!50!black},
	keywordstyle=\color{blue},
	stringstyle=\color{magenta},
	basicstyle=\linespread{1}\ttfamily,
	numberstyle=\tiny,
	breakatwhitespace=false,
	breaklines=true,
	captionpos=t,
	frame=single,
	keepspaces=true,
	language=matlab,
	numbers=none,
	numbersep=5pt,
	showspaces=false,
	showstringspaces=false,
	showtabs=false,
	tabsize=2,
	aboveskip=1em,
	belowskip=1em,
	belowcaptionskip=12pt
}

\usepackage{fancyhdr}
\newcommand{\myname}{Sean Rendell}
\newcommand{\mycourse}{MSc. Artifical Intelligence and Machine Learning}
\newcommand{\projecttitle}{Low Rank Groupwise Deformations for Motion Tracking in Cardiac Cine MRI}
\newcommand{\publishdate}{September 2023}

\DeclareMathOperator*{\argmin}{arg\,min}
\DeclareMathOperator{\vectorise}{vec}
\begin{document}

	\thispagestyle{empty}
	\begin{center}
		\vspace*{2cm}
		{\includegraphics[width = 12cm]{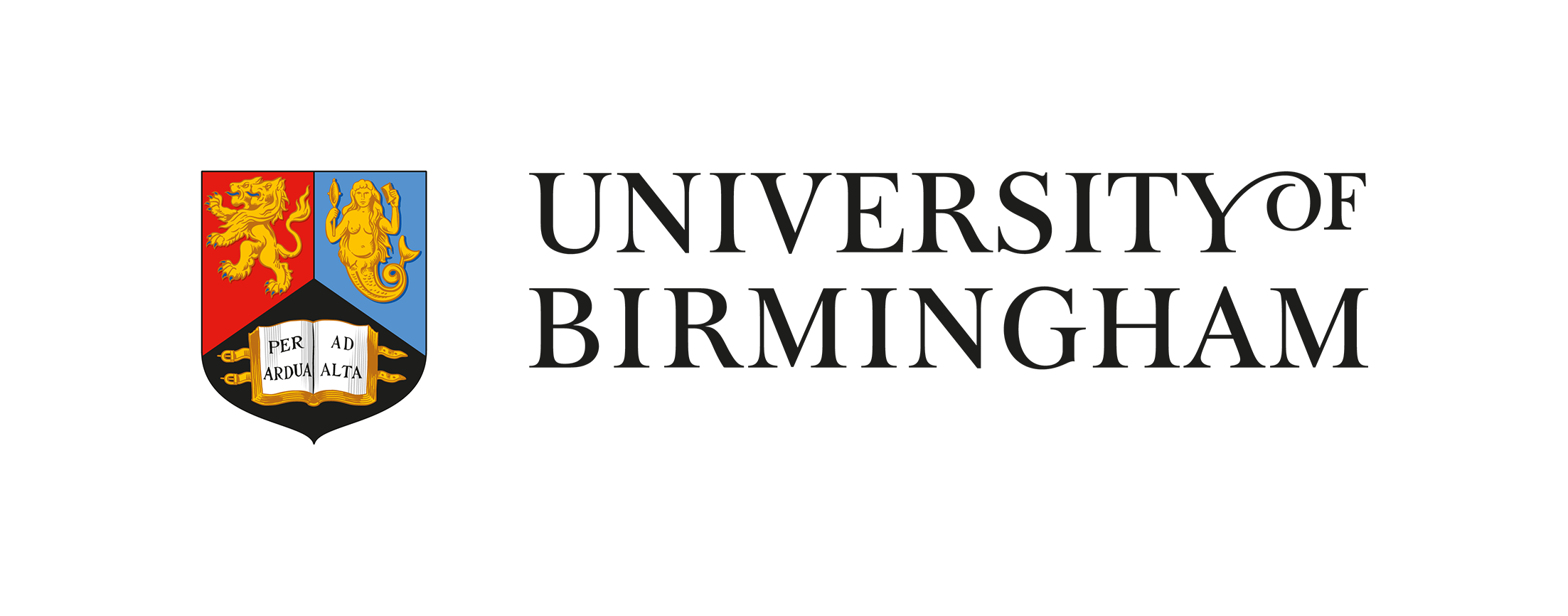} \par}
		\vspace{1cm}
		{\bfseries\Huge \myname \par}
		{\bfseries\Large \mycourse \par}
		\vspace{1cm}
		{\bfseries\Huge \projecttitle \par}
		\vspace{1cm}
		{\bfseries\Large \publishdate \par}
		\vfill
		{\bfseries\Large Department of Computer Science \par}
		\vspace{1cm}
	\end{center}

\pagenumbering{roman}
\setcounter{page}{0}

\chapter*{Abstract}
Diffeomorphic image registration is a commonly used method to deform one image to resemble another. While warping a single image to another is useful, it can be advantageous to warp multiple images simultaneously, such as in tracking the motion of the heart across a sequence of images. In this paper, our objective is to propose a novel method capable of registering a group or sequence of images to a target image, resulting in registered images that appear identical and therefore have a low rank. Moreover, we aim for these registered images to closely resemble the target image. Through experimental evidence, we will demonstrate our method's superior efficacy in producing low-rank groupwise deformations compared to other state-of-the-art approaches.

\chapter*{Acknowledgements}
I want to thank my supervisor Dr. Jinming Duan for his support and guidance during the course of this project.

\setcounter{tocdepth}{2}
\tableofcontents
\listoffigures
\listoftables

\newpage
\setcounter{page}{0}
\pagenumbering{arabic}

\newtheorem{example}{Example}[section]

\chapter{Introduction}
\label{cha:Introduction}

The following project was to try and take steps towards making an automatic framework for estimating cardiac strain in patients. In this chapter, we will briefly describe the background required to understand this task, then we will explain why this project is important and set out exactly what we aim to accomplish, and then describe what my project contributes to the literature. We will then explain some of the broader literature, not necessarily what we have focused on, but an overview of some related work in this general topic area. Finally, in detail, we will begin to explain my project with examples and analysis, finishing with a conclusion of what we have achieved. 

\section{Background}
\label{sec:Background}

Strain estimation is important in medicine as the strain on a patient's heart can be used to identify disorders or problems with the heart instead of using left ventricular ejection fraction (LVEF) as there is "growing evidence [to] suggest [strain] may improve our ability to risk stratify these groups by detecting systolic dysfunction despite a normal LVEF" \cite{ejvsstrain}. This is to say that instead of using LVEF, we can use cardiac strain to identify patients with valvular heart disease as these patients might show a normal LVEF. Furthermore, \cite{cardiacstrain} also states how strain can be used to identify patients with systolic dysfunction despite a normal LVEF. However, they also provide multiple other benefits to using cardiac strain too, for example, use for chemotherapy patients to "identify sub-clinical LV dysfunction". These papers demonstrate the importace and benefits that can be provided from being able to estimate cardiac strain and this means that if we can create an algorithm to automatically identify cardiac strain then this can be used clinically in the diagnosis of heart conditions. This could potentially save lives. While automatic strain estimation is very complex, we aim to create algorithms which could help get us and others more accurately estimate cardiac strain, and in the future we could add to these to estimate cardiac strain ourselves. Although we do not estimate strain, we aim to add to the foundations which strain estimation is built upon - this involves topics like diffeomorphic image registration, robust principle component analysis, and motion tracking. Diffeomorphic image registration is a process of doing a diffeomorphic (non-linear) mapping from one image to another; such that you can warp one image to look like identical to the other. Robust principle component analysis (RPCA) is a variation of principle component analysis (PCA) which is used to perform dimensionality reduction. It turns out that RPCA will be very useful for motion tracking. Motion tracking, in this context, will use diffeomorphic image registration to track the motion of an object throughout a sequence of images (instead of between a pair of images) by warping a sequence of images to look like a single image. \\

To aid us with this we make use of cardiac MRI datasets like the ACDC dataset \cite{dataset} or the BioBank dataset \cite{biobank}. These datasets provide us with hundreds of cardiac MRI videos along with ground truth segmentation labels of the sections of the heart that will help us test our algorithms and provide a thorough analysis.

\section{Motivations and Aims}
\label{sec:Motivations and Aims}

As stated in section \ref{sec:Background}, cardiac strain estimation can be used clinically for diagnosis and identification of heart conditions like valvular heart disease. Therefore, while we may not be able to estimate cardiac strain, we aim to take steps towards estimating cardiac strain. This is because in order to estimate cardiac strain one must be able to do multiple things first - like track the motion of the heart. This work is important as if we can help provide new approaches to help estimate cardiac strain then we can potentially help with diagnosis of cardiac conditions; the benefits of this are huge. For these reasons, we aim to provide a motion tracking algorithm which can be used to track the motion of the heart. We will achieve this by first making use of the current literature in diffeomorphic image registration to implement an algorithm in Pytorch that can perform diffeomorphic image registration. After this, we will use what we have done to take a new approach in order to track the motion of the heart. To achieve this goal, we will mathematically justify our approach, and then using that, in conjunction with Pytorch, we will test our new approach on sample data to evaluate our success. 

\section{Contributions}
\label{sec:Contributions}

We implement an approach for diffeomorphic image registration following the method set out in \cite{thorley2021nesterov}. We test this approach on a cardiac imaging dataset. This is implemented using Pytorch and a thorough analysis of results is included too. \\

We implement the code for a method called robust principle component analysis \cite{candes2009robust} and give examples of the output of the code using figures.\\

Furthermore, we combine the ideas of diffeomorphic image registration and robust principle component analysis to perform motion tracking on the same cardiac imaging dataset. While these two methods have been combined before \cite{haase2020deformable}, the approach we take to combine them is a new one that is implemented again using Pytorch. We provide an analysis of this new approach and provide detail explanations of the upsides and what's new that the algorithm can do; furthermore, we compare it to the diffeomorphic image registration model to be certain that it does perform better. 

\chapter{Related Work}
\label{cha:Related Work}

Other works can either tackle the whole problem that we aim to tackle or just part of the problem. If it is just part of the problem then we can still learn from what they do and help it to guide our approach at the whole thing. Lots of work around the topic of estimating cardiac motion utilise deep learning. For example, the work in \cite{chen2020anatomyaware} where they use a variational auto-encoder (VAE) to handle the task of estimating cardiac motion. Their VAE is used to impose a restriction on the shape of the myocardium so that the VAE is "aware" of the anatomy and thus it is better equipped to track the motion of the heart as it knows what shape it should be. They provide evidence of good results, however, the use of deep learning is very data driven and therefore requires good labelled data from professionals - this is difficult to obtain and can make deep learning in some situations unfeasible. A similar approach is shown in \cite{2019siameseconvolution} where they use a Siamese convolutional network which is essentially two convolutional networks which are in parallel and share a lot of weights. The input to this network will be two patches of the left ventricle at two time frames, and the output extracts features which contour the left ventricle. These contours can of course then be used to estimate the motion of the left ventricle. Again, this suffers from the same downside as \cite{chen2020anatomyaware} that it utilises deep learning which can be very computationally expensive to train. In \cite{2019siameseconvolution} they state how the training was done on a "Nvidia GTA 1080	Ti and 11 GB RAM" which may not be accessible in all circumstances. Deep learning can be very helpful as can be an asset though if the task requires a deep learning approach. However, the approach I will be using for diffeomorphic image registration and motion estimation will not be based off of deep learning methods. In \cite{LDDMM} they examine the large deformation diffeomorphic metric mapping problem (LDDMM) whereby assuming smooth deformations they minimise the vector field to perform diffeomorphic image registration numerically without the use of deep learning. Furthermore, in \cite{FLASH} they also perform numerical diffeomorphic image registration via gradient descent in a "low-dimensional approximated Lie algebras" which also performs diffeomorphisms with a time-varying velocity, however, this is an approximated low-dimensional velocity which greatly increases computation performance. In \cite{variablecontrast} they perform diffeomorphic image registration via a technique that does not rely upon matching image intensities. While they provide excellent results when compared against \cite{diffeodemons}, the technique they make use of is one we will not be using as we will find matching intensities very beneficial. These works highlight that diffeomorphic image registration does not need to be done via deep learning methods and that numerical methods, which we will be making use of, are perfect for the task due to the fact that they do not require huge amounts of training data. \\

In \cite{fechter2020shot} \cite{hering2018enhancing} they apply deep learning approaches for image registration. \cite{fechter2020shot} focuses on their applications for periodic motion tracking. This further shows that deep learning is effective for image registration, but more importantly, that this can be extended to track periodic motion which we are interested in as cardiac motion is of course periodic. In \cite{LDDMM}, \cite{FLASH} they only apply their methods for diffeomorphic image registration, they are not concerned with motion tracking. In \cite{pcalungmotion} they use a principle component analysis based approach to track the motion of the lung. This paper demonstrated very good results and showed promise that principle component analysis based approaches can be used for motion tracking. Some of the literature apply image registration followed by some other technique for accurate motion tracking. For example, in \cite{haase2020deformable} they perform diffeomorphic image registration first and then apply a principle component analysis (PCA) based approach called robust principle component analysis (RPCA) for motion tracking. They applied their method to real-world data and showed excellent results that further show that a PCA-based approach is viable, more importantly, that RPCA is an excellent approach for motion tracking. For motion tracking we will make use of a similar technique, as we've seen excellent results and seen that PCA based approaches for motion tracking work well. 

\chapter{Diffeomorphic Image Registration}
\label{cha:Diffeomorphic Image Registration}

Our main goal is to track motion, but before we do that we need to investigate diffeomorphic image registration. While the term might sound complex we can break it down to explain it simply. Image registration is the process of mapping one image onto another such that they look similar, traditionally image registration was done with matrix transformation such as translations, rotations and so on. Unfortunately, these transformation are not always the best approach as two images might have different sections moving in different directions. In this case linear matrix transformations will not be able to map one image onto another, to do this we need a non-linear approach and this is where the diffeomorphic image registration comes in. It is exactly the same, where you want to register one image onto another such that they look similar, except in this case the transformation is explained by a diffeomorphism. A diffeomorphism is a map between two images (or sets) of the form $$f:X \to Y$$ such that the map is bijective and the map and it's inverse are both differentiable. A bijective mapping between two sets means that the mapping is both surjective and injective. Surjective (onto) meaning every element in the second set, $Y$, is mapped onto. Injective (one-to-one) means that every element in the first set $X$ only maps onto one element in the set $Y$. This means that diffeomorphic image registration is a way to map one image onto another such that the mapping is a diffeomorphism. Before we go onto explaining how to calculate this diffeomorphism, we must first look at optical flow which will tell us where things are moving between a pair of images. This information will be useful as if we know where things are moving, then we will be able to calculate a mapping for every element in the first image onto every element in the second image. Therefore, once we have calculated the optical flow between a pair of images we can then move on to performing diffeomorphic image registration to map one image onto another. 

\section{Optical Flow}
\label{sec:Optical Flow}

If we wanted to approximate the motion of pixels between two consecutive images we would want to use what's called optical flow. Optical flow is used to approximate the motion from one image to another and the assumption it is based on is that we assume the intensity of a pixel in one image is also present in the next image. For instance, if we had a 3D image (2D + temporal dimension) and a pixel from that image with intensity $I(x,y,t)$ at co-ordinates $(x,y,t)$, then we should assume in the next image at time-step $t+\Delta t$ that same intensity $I$ should also be in it. However, the co-ordinates of that intensity may not be in the same location, it may how moved by some small margin, $\Delta$. The new co-ordinates may be $(x+ \Delta x, y + \Delta y, t + \Delta t)$, which, assuming the intensity is the same, has intensity $I(x+ \Delta x, y + \Delta y, t + \Delta t)$. Since we are assuming these intensities appear in both images, we can equate them and so we get the brightness constancy equation:

\begin{align}
	\label{eq:Brightness Constancy}
	I(x,y,t) = I(x+\Delta x, y+\Delta y, t+\Delta t)
\end{align}

such that the intensities in the first image (time step $t$) are equal to the intensities in the second image (time step $\Delta t$) that have moved by a small amount $\Delta x$ or $\Delta y$. We can expand the right hand side of Equation \ref{eq:Brightness Constancy} by using a Taylor series expansion where we assume the change is small such that:

\begin{align}
	\label{eq:Brightness Constancy Taylor}
	I(x+\Delta x, y+\Delta y, t+\Delta t) = I(x,y,t) + \frac{\partial I}{\partial x}\Delta x + \frac{\partial I}{\partial y}\Delta y + \frac{\partial I}{\partial t}\Delta t + ...
\end{align}

For these to be equal, the Taylor Series Expansion must have infinitely many terms, however, that is infeasible. Instead of this we make Equation \ref{eq:Brightness Constancy Taylor} an approximation by truncating terms with a partial derivative of order greater than 1. Not only does this make the equations easier to solve, it also speeds up the computation process, however, it compromises on some accuracy. Since in Equation \ref{eq:Brightness Constancy} with truncation we have that $I(x+\Delta x, y+\Delta y, t+\Delta t) = I(x,y,t)$ but in Equation \ref{eq:Brightness Constancy Taylor} we have that $I(x+\Delta x, y+\Delta y, t+\Delta t) \approx I(x,y,t) + \frac{\partial I}{\partial x}\Delta x + \frac{\partial I}{\partial y}\Delta y + \frac{\partial I}{\partial t}\Delta t$ we see that for Equation \ref{eq:Brightness Constancy} to be true, the sum of the partial derivatives must be equal to $0$. This gives us:

\begin{align}
	\label{eq:Optical Flow}
	\frac{\partial I}{\partial x}\Delta x + \frac{\partial I}{\partial y}\Delta y + \frac{\partial I}{\partial t}\Delta t &= 0 \\
	\label{eq:Optical Flow 2}
	\implies \frac{\partial I}{\partial x}\frac{\Delta x}{\Delta t} + \frac{\partial I}{\partial y}\frac{\Delta y}{\Delta t} + \frac{\partial I}{\partial t}\frac{\Delta t}{\Delta t} &= 0 \\
	\label{eq:Optical Flow 3}
	= \frac{\partial I}{\partial x}V_x + \frac{\partial I}{\partial y}V_y + \frac{\partial I}{\partial t} &= 0
\end{align}

where $V_x$ and $V_y$ are the velocities of optical flow in the image, and $\frac{\partial I}{\partial x}$, $\frac{\partial I}{\partial y}$, $\frac{\partial I}{\partial t}$ are the derivatives of the image in each dimension. Equation \ref{eq:Optical Flow 3} has 2 unknown variables (remember we wish to determine the flow $V_x$ and $V_y$) but only one equation and is therefore an underdetermined system and therefore has no solutions. \\

One proposed method of solving this was done by Lucas and Kanade \cite{lucaskanade} where we assume the motion between two consecutive images is small within a small given window. This gives us a system of equations of the form:

\begin{align}
	\label{eq:Lucas Kanade}
	\frac{\partial I(q_1)}{\partial x}V_x + \frac{\partial I(q_1)}{\partial y}V_y &= -\frac{\partial I(q_1)}{\partial t} \\
	\frac{\partial I(q_2)}{\partial x}V_x + \frac{\partial I(q_2)}{\partial y}V_y &= -\frac{\partial I(q_2)}{\partial t} \\
	&\vdots \\
	\frac{\partial I(q_n)}{\partial x}V_x + \frac{\partial I(q_n)}{\partial y}V_y &= -\frac{\partial I(q_n)}{\partial t}
\end{align}

where $\frac{\partial I(q_n)}{\partial x}$ denotes the partial derivative of $I$ with respect to $x$ when evaluated at position $q_n = (x_n, y_n)$ and $q_n$ are pixels within a small window. From this we can transform equation \ref{eq:Lucas Kanade} into vector notation as $\vec{A}\vec{v} = \vec{b}$ such that:

\begin{align*}
	\vec{A} = \begin{bmatrix}
		\frac{\partial I(q_1)}{\partial x} & \frac{\partial I(q_1)}{\partial y} \\
		\frac{\partial I(q_2)}{\partial x} & \frac{\partial I(q_2)}{\partial y} \\
		\vdots & \vdots \\
		\frac{\partial I(q_n)}{\partial x} & \frac{\partial I(q_n)}{\partial y}
	\end{bmatrix}, \vec{v} = \begin{bmatrix}
	V_x \\
	V_y
\end{bmatrix}, \vec{b} = \begin{bmatrix}
-\frac{\partial I(q_1)}{\partial t} \\
-\frac{\partial I(q_2)}{\partial t} \\
\vdots \\
-\frac{\partial I(q_n)}{\partial t}
\end{bmatrix}
\end{align*}

Now we have the equation $\vec{A}\vec{v} = \vec{b}$ we would wish to solve this for $\vec{v}$, ideally we would just take the left inverse of $\vec{A}$ on both sides but we cannot do this as $\vec{A}$ is not square. To circumvent this problem we can apply the transpose of $\vec{A}$ to the left of both sides of the equation giving us: $\vec{A}^T\vec{A}\vec{v} = \vec{A}^T\vec{b}$. Clearly, $\vec{A}^T\vec{A}$ is a square matrix and so we can take it's inverse giving us a final solution for $\vec{v}$ as:

\begin{align}
	\label{eq:Lucas Kanade Solution}
	\vec{v} = (\vec{A}^T\vec{A})^{-1}\vec{A}^T\vec{b}
\end{align}

Equation \ref{eq:Lucas Kanade Solution} gives the optical flow between a pair of images, however, it can only recover the velocity if $\vec{A}^T\vec{A}$ is invertible, and also if the flow is small; it cannot recover large flows. Furthermore, due to the assumption of the pixels being within a window it cannot recover the flow beyond the proposed region \cite{lucaskanadeanalysis}. Since using the Lucas-Kanade method of solving optical flow has too many downsides; we must look for other methods. \\

Another method for solving the optical flow Equation \ref{eq:Optical Flow 3} is to introduce a regularisation term imposing smoothness. This means that the solution to the optical flow equation prefers flows that are small and thus will be smooth. This method was proposed by Horn and Schunck in \cite{horn_schunck} and this transforms the optical flow equation into the following minimisation problem:

\begin{align}
	\label{eq:Horn Schunck}
	\min_\vec{v}\{\frac{1}{2}\|\frac{\partial I_1}{\partial x}V_x + \frac{\partial I_1}{\partial y}V_y + I_1 - I_0\|^2 + \frac{\lambda}{2}\|\nabla \vec{v}\|^2\} 
\end{align}

where $I_1$ is the image at time step $t + \Delta t$ and $I_0$ is the image at time step $t$. Thus we get that $I_1 - I_0 = \frac{\partial I}{\partial t}$. We change the notation as it is more readable and see that $\vec{v} = [V_x, V_y]^{T}$. Furthermore, $\nabla$ is the gradient operator. $\lambda$ is the regularisation parameter controlling how smooth the deformations will be, a high value for $\lambda$ will mean we have smooth deformations. In Equation \ref{eq:Horn Schunck}, the left hand term is the same as the equation in Equation \ref{eq:Optical Flow 3}, however, we have introduced the right hand term in Equation \ref{eq:Horn Schunck}, which is the regularisation term also known as a total variation regularisation. Solving the minimisation problem in Equation \ref{eq:Horn Schunck} would not give us a sufficient solution as the velocities will be too small and we would not register one image onto another, particularly if the deformation from the source image to the target image is large. To solve this we incorporate ideas from diffeomorphic image registration. 

\section{Diffeomorphic Image Registration}
\label{sec:Diffeomorphic Image Registration}

The paper \cite{fastdiffeoIRalgorithm} defined a diffeomorphic deformation by the ODE:

\begin{align}
	\label{eq:Diffeomorphic IR}
	\frac{\partial \phi}{\partial t} = \vec{v}_t(\phi_t) : t\in[0,1]
\end{align}

where $\phi_t$ is a mapping of coordinates from one image to another and $\phi_0 = Id$ where $Id$ is the identity mapping (mapping an image onto itself). A simple method for solving this ODE is with Euler integration as shown in many works like \cite{thorley2021nesterov}\cite{diffeoIReulerexample} where the final deformation is calculated using a composition of small deformations. If we have $N$ uniformly spaced time-steps, $\phi_1$ can be calculated as:

\begin{align}
	\label{eq:Diffeomorphic IR phi1}
	\phi_1 = (Id + \frac{\vec{v}_{t_{N-1}}}{N}) \circ (Id + \frac{\vec{v}_{t_{N-2}}}{N}) \circ ... \circ (Id + \frac{\vec{v}_{t_{1}}}{N}) \circ (Id + \frac{\vec{v}_{t_{0}}}{N})
\end{align}

where $\circ$ denotes the composition operator. The method for calculating the final deformation field that we will be using in this paper will be the method set out in \cite{thorley2021nesterov}. For a pair of images $I_1$ and $I_0$ being the source and target images respectively, first we solve the Horn Schunck method for calculating optical flow, Equation \ref{eq:Horn Schunck}, to get $\vec{v}^{*}_{t_{N-1}}$. The source image $I_1$ is then warped to $I_1^{\omega}$ by: $I_1^{\omega} = I_1 \circ (Id + \vec{v}^{*}_{t_{N-1}})$ where the denominator $N$ is set equal to $1$. Then we solve the Horn Schunck equation again except this time by using $I_1^{\omega}$ and $I_0$ which will give us $\vec{v}^{*}_{t_{N-2}}$. Next we update the warped source $I_1^{\omega}$ to $I_1^{\omega}$ by: $I_1^{\omega} = I_1 \circ (Id + \vec{v}^{*}_{t_{N-1}}) \circ (Id + \vec{v}^{*}_{t_{N-2}})$. We repeat these steps of solving the Horn Schunck equation using $I_1^{\omega}$ and $I_0$ and using the output to update $I_1^{\omega}$. This gives us our general solution of the form:

\begin{align}
	\label{eq:Warped Source Solution}
	I_1^{\omega} = I_1 \circ (Id + \vec{v}^{*}_{t_{N-1}}) \circ (Id + \vec{v}^{*}_{t_{N-2}}) \circ ... \circ (Id + \vec{v}^{*}_{t_{1}}) \circ (Id + \vec{v}^{*}_{t_{0}})
\end{align}

where $\vec{v}^{*}_{t}$ is a solution to the Horn Schunck Equation \ref{eq:Horn Schunck}. It is noted in \cite{thorley2021nesterov} that "the first-order diffusion regularisation performs poorly against large deformations" and "To circumvent these shortcomings of the HS model, we propose a new, general variational model, given by:"

\begin{align}
	\label{eq:Variational HS}
	\min_\vec{v}\{\frac{1}{2}\|\rho(\vec{v})\|^{2} + \frac{\lambda}{2}\|\nabla^{n}\vec{v}\|^{2}\}
\end{align}

where $\rho(\vec{v}) = \frac{\partial I_1}{\partial x}V_x + \frac{\partial I_1}{\partial y}V_y + I_1 - I_0$, $\vec{v} = [V_x, V_y]^T$ and $n>0$. In \cite{thorley2021nesterov}, the first term of equation \ref{eq:Variational HS} replaced the $2$ with $s \in \{1,2\}$, however, in our paper we only consider the case where $s=2$ and as such we have removed it and replaced it with $2$. Furthermore, $\nabla^{n}$ is known as the arbitrary order gradient, where for a function $f(x,y)$, $\nabla^n f = [\binom{n}{k_1, k_2}\frac{\partial^n f}{\partial x^{k_1} \partial y^{k_2}}]^T$ where $k_i = 0,...,n$, $i \in \{1,2\}$ and $k_1 + k_2 = n$. Finally $\binom{n}{k_1, k_2}$ are known as binomial coefficients and are calculated by $\frac{n!}{k_1! k_2!}$ \\

To solve this, \cite{thorley2021nesterov} proposed using alternating direction method of multipliers (ADMM) \cite{admm}. However, they used ADMM accelerated by Nesterov's method \cite{nesterovaccelerated}. Although I will be using the method set out in \cite{thorley2021nesterov}, we will not be using Nesterov accelerated ADMM, we will instead be using the over-relaxed method that can be seen in \cite{arbitraryordertotalvariation}. We will declare how this makes the method we used different to the method used in \cite{thorley2021nesterov} soon. \\

The first step in solving Equation \ref{eq:Variational HS} is to decouple the data term and the regularisation term as so:

\begin{align}
	\label{eq:Decoupled Variational HS}
	\argmin_\vec{v}\{\frac{1}{2}\|\rho(\vec{v})\|^{2} + \frac{\lambda}{2}\|\nabla^{n}\vec{w}\|^{2}\} : \vec{w} = \vec{v}
\end{align}

this will allow a closed-form solution to be found with respect to both variables. This also turns the problem into a constrained optimisation problem and you can solve constrained optimisation problems using Lagrange multipliers. Using this gives us the Lagrangian parametrised by $\vec{b}$:

\begin{align}
	\label{eq:Lagrangian HS}
	\mathcal{L}_\mathcal{A}(\vec{v},\vec{w};\vec{b})=\frac{1}{2}\|\rho(\vec{v})\|^{2} + \frac{\lambda}{2}\|\nabla^{n}\vec{w}\|^{2} + \frac{\theta}{2}\|\vec{w} - \vec{v} - \vec{b}\|^2
\end{align}

where $\vec{b}$ is a Lagrangian multiplier and $\theta>0$ is a penalty parameter which has an effect on convergence. The problem in Equation \ref{eq:Lagrangian HS} can be broken down into subproblems to solve for $\vec{v}$, $\vec{w}$ and $\vec{b}$ by considering the terms which depend on each of these respectively and considering how to solve for $\vec{b}$ in \cite{admm}. This gives us the following set of equations using the over-relaxed solver in \cite{arbitraryordertotalvariation}, instead of the original Nesterov accelerated method as:

\begin{align}
	\label{eq:Lagrangian Expanded}
	\begin{cases}
		\vec{v}^k = \argmin_\vec{v} \frac{1}{2}\|\rho(\vec{v})\|^2 + \frac{\theta}{2}\|\vec{w}^k -\vec{v}^{k-1} -\vec{b}^k\|^2 \\
		\hat{\vec{v}}^k = \alpha\vec{v}^k + (1-\alpha)\vec{w}^{k-1}\\
		\vec{w}^k = \argmin_\vec{w} \frac{\lambda}{2}\|\nabla^n \vec{w}^{k-1}\|^2 + \frac{\theta}{2}\|\vec{w}^{k-1} - \hat{\vec{v}}^{k} -\vec{b}^{k-1}\|^2 \\
		\vec{b}^k = \vec{b}^{k-1} + \hat{\vec{v}}^k - \vec{w}^k\\
	\end{cases}
\end{align}

At the beginning of our iterations we set $\vec{v} = 0, \vec{w} = 0$ and $\vec{b} = 0$. Due to the decoupling that was done in equation \ref{eq:Decoupled Variational HS} a closed form solution can be derived for both $\vec{v}$ and $\vec{w}$. The closed form solution for $\vec{v}$ is derived as follows:

\begin{align*}
	&\vec{v}^k = \argmin_\vec{v} \frac{1}{2}\|\rho(\vec{v})\|^2 + \frac{\theta}{2}\|\vec{w}^k -\vec{v} -\vec{b}^k\|^2 \\
	&\frac{\partial\rho(\vec{v})}{\partial\vec{v}}\rho(\vec{v}) - \theta(\vec{w}-\vec{v}-\vec{b})\frac{\partial\vec{v}}{\partial\vec{v}} = 0\\
	&\vec{J}\left(\frac{\partial I_1}{\partial x}V_x + \frac{\partial I_1}{\partial y}V_y + I_1 - I_0\right) + \theta\vec{v} = \theta(\vec{w}-\vec{b})\\
	&[\vec{J}\vec{J}^T + \theta\mathbb{1}]\vec{v} + \vec{J}(I_1 - I_0) = \theta(\vec{w}-\vec{b})\\
	&[\vec{J}\vec{J}^T + \theta\mathbb{1}]\vec{v} = \theta(\vec{w}-\vec{b}) - \vec{J}(I_1 - I_0)\\
\end{align*}
Where $\vec{J} = [\frac{\partial I_1}{\partial x},                                                                                                                                                                                                                                                                      \frac{\partial I_1}{\partial y}]^T$ and $\mathbb{1}$ is the identity matrix. Due to the identity being invertible we can use the Sherman-Morrison equation to calculate to inverse matrix and derive a final closed form solution for $\vec{v}$ \cite{diffeodemons}. To do this all we need to do is calculate the inverse of $(\theta \mathbb{1} + \vec{J}\vec{J}^T)$

\begin{align}
	\label{eq:Sherman-Morrison}
	(A + uv^T)^{-1} &= A^{-1} - \frac{A^{-1}uv^TA^{-1}}{1 + v^TA^{-1}u}\\
	(\theta \mathbb{1} + \vec{J}\vec{J}^T)^{-1} &= (\theta \mathbb{1} )^{-1} - \frac{(\theta \mathbb{1} )^{-1}\vec{J}\vec{J}^T(\theta \mathbb{1} )^{-1}}{1 + \vec{J}^T(\theta \mathbb{1} )^{-1}\vec{J}}\nonumber\\
	&= \frac{\mathbb{1}}{\theta} - \frac{\frac{\mathbb{1}}{\theta}\vec{J}\vec{J}^T\frac{\mathbb{1}}{\theta}}{1 + \vec{J}^T\frac{\mathbb{1}}{\theta}\vec{J}}\nonumber\\
	&= \frac{\mathbb{1}}{\theta} - \frac{\frac{\vec{J}\vec{J}^T}{\theta^2}}{1 + \frac{\vec{J}^T\vec{J}}{\theta}}\nonumber\\
	&= \frac{\mathbb{1}}{\theta} - \frac{\frac{\vec{J}\vec{J}^T}{\theta^2}}{\frac{\theta + \vec{J}^T\vec{J}}{\theta}}\nonumber\\
	&= \frac{\mathbb{1}}{\theta} - \frac{\vec{J}\vec{J}^T}{\theta(\theta + \vec{J}^T\vec{J})}\nonumber\\
	\label{eq:Sherman-Morrison 1}
	&= \frac{(\mathbb{1}\theta + \mathbb{1}\vec{J}^T\vec{J}) - \vec{J}\vec{J}^T}{\theta(\theta + \vec{J}^T\vec{J})}\\
	&= \frac{\begin{bmatrix}
			\biggl(\frac{\partial I}{\partial x}\biggr)^2 + \biggl(\frac{\partial I}{\partial y}\biggr)^2 + \theta & 0 \\
			0 & \biggl(\frac{\partial I}{\partial x}\biggr)^2 + \biggl(\frac{\partial I}{\partial y}\biggr)^2 + \theta
		\end{bmatrix} - \begin{bmatrix}
			\biggl(\frac{\partial I}{\partial x}\biggr)^2 & \frac{\partial I}{\partial x}\frac{\partial I}{\partial y} \nonumber\\
			\frac{\partial I}{\partial y}\frac{\partial I}{\partial x} & \biggl(\frac{\partial I}{\partial y}\biggr)^2
	\end{bmatrix}}{\theta\biggl(\biggl(\frac{\partial I}{\partial x}\biggr)^2 + \biggl(\frac{\partial I}{\partial y}\biggr)^2 + \theta\biggr)}\nonumber\\
	\label{eq:Sherman-Morrison 2}
	&=\frac{\begin{bmatrix}
			\biggl(\biggl(\frac{\partial I}{\partial y}\biggr)^2 + \theta\biggr) & -\biggl(\frac{\partial I}{\partial x}\frac{\partial I}{\partial y}\biggr) \\
			-\biggl(\frac{\partial I}{\partial y}\frac{\partial I}{\partial x}\biggr) & \biggl(\biggl(\frac{\partial I}{\partial x}\biggr)^2 + \theta\biggr)
	\end{bmatrix}}{\theta\biggl(\biggl(\frac{\partial I}{\partial x}\biggr)^2 + \biggl(\frac{\partial I}{\partial y}\biggr)^2 + \theta\biggr)}
\end{align}

If we take the numerator of Equation \ref{eq:Sherman-Morrison 1} and multiply it by $\vec{J}$ we get:

\begin{align}
	\label{eq:Sherman-Morrison J equation}
	((\mathbb{1}\theta + \mathbb{1}\vec{J}^T\vec{J}) - \vec{J}\vec{J}^T)\vec{J} = \theta \vec{J}
\end{align}

Now that we have calculated the inverse of $(\theta \mathbb{1} + \vec{J}\vec{J}^T)$, which corresponds to Equation \ref{eq:Sherman-Morrison 2}, we can use this along with the identity calculated in Equation \ref{eq:Sherman-Morrison J equation} to calculate the closed form solution for $\vec{v}$, shown in Equation \ref{eq:Solved Lagrangian cases v}. The closed form solution for $\vec{w}$ is derived as follows:

\begin{align}
	&\vec{w}^k = \argmin_\vec{w} \frac{\lambda}{2}\|\nabla^n \vec{w}\|^2 + \frac{\theta}{2}\|\vec{w}^k - \hat{\vec{v}} -\vec{b}^k\|^2 \nonumber\\
	&\lambda\nabla^{2n}\vec{w} + \theta(\vec{w} - \vec{v} - \vec{b}) = 0\nonumber\\
	&\mathcal{F}(\lambda\nabla^{2n}\vec{w} + \theta(\vec{w} - \vec{v} - \vec{b})) = 0\nonumber\\
	&\lambda\mathcal{F}(\nabla^{2n}\vec{w}) + \theta(\mathcal{F}(\vec{w}) - \mathcal{F}(\vec{v}) - \mathcal{F}(\vec{b})) = 0\nonumber \\
	&\mathcal{F}(\vec{w})[\lambda\mathcal{F}(\Delta^n) + \theta] = \theta\mathcal{F}(\vec{v} + \vec{b}) \nonumber\\
	\label{eq:Solved Lagrangian w derivation}
	&\vec{w} = \mathcal{F}^{-1}\left(\frac{\theta\mathcal{F}(\hat{\vec{v}}^k+\vec{b}^k)}{\lambda\mathcal{F}(\Delta^n) + \theta}\right)
\end{align}

Where $\mathcal{F}$ is the discrete Fourier transform (DFT) which is used due to the periodic boundary conditions, this is important as \cite{thorley2021nesterov} uses the DFT whereas \cite{arbitraryordertotalvariation} uses the discrete cosine transform due to the Neumann boundary conditions that are used in that paper. The boundary conditions are important as they are used in discretisation where we go from continuous partial derivatives to discrete values. This is seen as we defined: $\nabla^n f = [\binom{n}{k_1, k_2}\frac{\partial^n f}{\partial x^{k_1} \partial y^{k_2}}]^T$ however we transform to $\mathcal{F}(\Delta^n)$. $\Delta^n$ is the $n$-th order Laplace operator which is discretised by finite difference. The solution to what $\mathcal{F}(\Delta^n)$ is equal to when discretised by finite difference is given in the following:

\begin{align}
	\label{eq:Fourier Transform Finite Diff Derivation}
	&\Delta u_{i,j} = -(u_{i+1,j} + u_{i-1,j} + u_{i,j+1} + u_{i,j-1} - 4u_{i,j})\\
	&\text{By the definition of a multidimensional discrete Fourier transform we get that:}\nonumber\\
	\label{eq:Fourier Transform Finite Diff Derivation 1}
	&\mathcal{F}(u_{i,j}) = \sum_{n=0}^{N-1}\sum_{m=0}^{M-1}\exp^{-\frac{i2\pi kn}{N}}\exp^{-\frac{i2\pi km}{M}} u_{i,j}\\
	&\text{By the shift theorem of the DFT we get the following 2 equations:}\nonumber\\
	\label{eq:Fourier Transform Finite Diff Derivation 2}
	&\mathcal{F}(u_{i\pm1,j}) = \exp^{\pm\frac{i2\pi p}{N}}\mathcal{F}(u_{i,j})\\
	\label{eq:Fourier Transform Finite Diff Derivation 3}
	&\mathcal{F}(u_{i,j\pm1}) = \exp^{\pm\frac{i2\pi q}{M}}\mathcal{F}(u_{i,j})\\
	&\text{By inserting equations \ref{eq:Fourier Transform Finite Diff Derivation 1}, \ref{eq:Fourier Transform Finite Diff Derivation 3} and \ref{eq:Fourier Transform Finite Diff Derivation 2} into \ref{eq:Fourier Transform Finite Diff Derivation} we get:}\nonumber\\
	\begin{split}
		&\mathcal{F}(\Delta u_{i,j}) = -\biggl[\exp^{\frac{i2\pi p}{N}}\mathcal{F}(u_{i,j}) + \exp^{-\frac{i2\pi p}{N}}\mathcal{F}(u_{i,j}) + \exp^{\frac{i2\pi q}{M}}\mathcal{F}(u_{i,j})\\
		&\qquad\qquad\qquad + \exp^{-\frac{i2\pi q}{M}}\mathcal{F}(u_{i,j}) - 4\mathcal{F}(u_{i,j})\biggr]\nonumber\\
	\end{split}
	\\
	&\mathcal{F}(\Delta u_{i,j}) = -\left(\exp^{\frac{i2\pi p}{N}} + \exp^{-\frac{i2\pi p}{N}} + \exp^{\frac{i2\pi q}{M}} + \exp^{-\frac{i2\pi q}{M}} - 4\right)\mathcal{F}(u_{i,j})\nonumber\\
	\begin{split}
		&\mathcal{F}(\Delta u_{i,j}) = -\biggl[\cos(\frac{2\pi p}{N}) + i\sin(\frac{2\pi p}{N}) + \cos(\frac{2\pi p}{N}) - i\sin(\frac{2\pi p}{N}) + \cos(\frac{2\pi q}{M}) + i\sin(\frac{2\pi q}{M})\\
		&\qquad\qquad\qquad + \cos(\frac{2\pi q}{M}) - i\sin(\frac{2\pi q}{M}) - 4\biggr]\mathcal{F}(u_{i,j})\nonumber\\
	\end{split}
	\\
	&\mathcal{F}(\Delta) = -\left(2\cos(\frac{2\pi p}{N}) + 2\cos(\frac{2\pi q}{M}) - 4\right)\nonumber\\
	&\mathcal{F}(\Delta) = 2\left(2 - \cos(\frac{2\pi p}{N}) - \cos(\frac{2\pi q}{M})\right)
\end{align}

Subsequently we have that, $\mathcal{F}(\Delta^n) = 2^n\left(2 - \cos(\frac{2\pi p}{M}) - \cos(\frac{2\pi q}{N})\right)^n$ where $p \in [0, M)$ and $q \in [0, N)$ and $p,q$ are grid indices. 

After all this, using Equations \ref{eq:Solved Lagrangian w derivation}, \ref{eq:Sherman-Morrison 2} and \ref{eq:Sherman-Morrison J equation} we get the final solutions to Equation \ref{eq:Lagrangian HS} as:

	\begin{numcases}{}
		\label{eq:Solved Lagrangian cases v}
		\vec{v}^k = \frac{\begin{bmatrix}
				\biggl(\biggl(\frac{\partial I}{\partial y}\biggr)^2 + \theta\biggr) & -\biggl(\frac{\partial I}{\partial x}\frac{\partial I}{\partial y}\biggr) \\
				-\biggl(\frac{\partial I}{\partial y}\frac{\partial I}{\partial x}\biggr) & \biggl(\biggl(\frac{\partial I}{\partial x}\biggr)^2 + \theta\biggr)
			\end{bmatrix}(\vec{w}^{k-1}-\vec{b}^{k-1}) - \vec{J}(I_1 - I_0)}{\biggl(\biggl(\frac{\partial I}{\partial x}\biggr)^2 + \biggl(\frac{\partial I}{\partial y}\biggr)^2 + \theta\biggr)}\\
		\label{eq:Solved Lagrangian cases v hat}
		\hat{\vec{v}}^k = \alpha\vec{v}^k + (1-\alpha)\vec{w}^{k-1}\\
		\label{eq:Solved Lagrangian cases w}
		\vec{w}^k = \mathcal{F}^{-1}\left(\frac{\theta\mathcal{F}(\hat{\vec{v}}^k+\vec{b}^{k-1})}{\lambda\mathcal{F}(\Delta^n) + \theta}\right)\\
		\label{eq:Solved Lagrangian cases b}
		\vec{b}^k = \vec{b} + \hat{\vec{v}}^k - \vec{w}^k
	\end{numcases}
With these closed-form solutions calculated we can now move onto the algorithm for performing diffeomorphic image registration.

\begin{algorithm}
	\caption{Diffeomorphic Image Registration} \label{alg:Diffeomorphic Image Registration}
	\textbf{Input Images:} $I_0$ and $I_1$\\
	\textbf{Input Parameters:} (levels, $N_{warp}$, $N_{iter}$, $\lambda$, $\theta$, $\alpha$, tolerance, difference)\\
	\For{\normalfont{\textbf{all}} $s \in$ levels}{
		$I_0 \gets resize^-(I_0, s)$\\
		$I_1 \gets resize^-(I_1, s)$\\
		\eIf{\normalfont$s = $ levels[$0$]}{
		$\vec{v}^0 = \vec{0}$
		}{
		$\vec{v}^s \gets resize^+(\vec{v}^s, 2)$
		}
		\While{\normalfont$\omega < N_{warp}$ or while Equation \ref{eq:Stopping Criteria Warp} is not true}{
		\If{$\omega = 0$}{
		$\vec{v}^{\omega} \gets \vec{v}^s$
		}
		$I_1^{\omega} \gets warp(I_1, \vec{v}^{\omega})$\\
		\textbf{Initialise:} $\vec{w}^{-1} = \vec{w}^0 = \vec{0}, \vec{b}^{-1} = \vec{b}^0 = \vec{0}$\\
			\While{\normalfont $k < N_{iter}$ or while Equation \ref{eq:Stopping Criteria Iter} is not true}{
			Update $\vec{v}^k$ with Equation \ref{eq:Solved Lagrangian cases v} and with $I_1^{\omega}$\\
			Update $\hat{\vec{v}}^k$ with Equation \ref{eq:Solved Lagrangian cases v hat}\\
			Update $\vec{w}^k$ with Equation \ref{eq:Solved Lagrangian cases w}\\
			Update $\vec{b}^k$ with Equation \ref{eq:Solved Lagrangian cases b}
			}
		$\vec{v}^{\omega} \gets \vec{v}^k$
		}
	$\vec{v}^s$ $\gets$ $\vec{v}^{\omega}$
	}
\textbf{Return:} $\vec{v}^* \gets \vec{v}^s$

\end{algorithm}

Algorithm \ref{alg:Diffeomorphic Image Registration} is the algorithm that was used for Diffeomorphic Image Registration in this paper. For clarification, we will now go through each step of Algorithm \ref{alg:Diffeomorphic Image Registration} and go into detail of how it works and why it is done. \\

\textbf{Initialisation}\\

First we input two images; $I_1$, the source image, and $I_0$, the target image. This algorithm will warp $I_1$ to look like $I_0$. Then we input the parameters:\\
levels - this parameter determines the scale factors that are used to decrease the size of the image. In this paper, $[4,2,1]$ is used for the scales. Note that scales must be powers of $2$ and must be in descending order. This is done so that the registration does not converge to local minima and overall increases the quality of the registration.\\
$N_{warp}$ and $N_{iter}$ control the number of iteration completed within the while loops. It is important to note that Algorithm \ref{alg:Diffeomorphic Image Registration} does include stopping criteria for the \textit{while} loops that are used. This means that we will complete the maximum number of iterations unless the stopping criteria is met in which case we will break the loop and move on. We used $N_{warp} = 20$ and $N_{iter}=1000$. \\
$\lambda$ first appears in Equation \ref{eq:Decoupled Variational HS} and so it is used to control regularisation. This was set to $\lambda=1$\\
$\theta$ is the penalty parameter from Equation \ref{eq:Lagrangian HS} and has an effect on the convergence, this was set to $\theta=0.01$, finally, we set $\alpha = 1.8$, $\alpha$ is called the relaxation parameter and setting the parameter above $1$ is what makes this method over-relaxation.\\
"tolerance" and "difference" are two parameters that are used to help guide the stopping criteria, these will be spoken about later when we speak about the stopping criteria themselves. \\

\textbf{Algorithm}\\

Next we loop over each value within the levels parameter. Then we have $\textit{resize}^-$ where the power of "-" denotes that the output will have a smaller dimension, so we are descaling $I_0$ and $I_1$ by a scale factor of $s$ where $s$ is a value within the levels parameter. \\

Next we have an if statement where if $s$ is the first value within the levels parameter then we initialise $\vec{v}$ which is what we wish to solve for. If $s$ is not the first value within the levels parameter then this loop has been done once before and therefore in the next stage the images will be twice as big (remember the levels parameter has values in powers of $2$) hence we must resize our velocity, $\vec{v}$, by a scale factor of $2$. (We are doing $\textit{resize}^+$ where in this case the "+" denotes that we are increasing the size of the output in regard to the input). \\

Next we have two while loops inside of each other. The first while loop runs with $\omega < N_{warp}$ and this happens for each scale. At the start of the loop we warp the source image $I_1$ to $I_1^{\omega}$ using the value of $\vec{v}^{\omega}$ to do the warping. After that, we use the warped source image for the next while loop. This loop goes while $k < N_{iter}$, and during this time we updates the parameters for $\vec{v}^k$, $\hat{\vec{v}}^k$, $\vec{w}^k$ and $\vec{b}^k$ using Equations \ref{eq:Solved Lagrangian cases v}, \ref{eq:Solved Lagrangian cases v hat}, \ref{eq:Solved Lagrangian cases w}, \ref{eq:Solved Lagrangian cases b} respectively. For both of these while loops they can both be ended pre-emptively if a stopping criteria is met. These will be listed below. Once the "inside" while loop is done, we update the value for $\vec{v}^{\omega}$ and then do the "inside" loop again. Once the "outer" while loop is completed we update the value for $\vec{v}^s$ and then repeat this process at the next scale. Once all scales have been completed, then we exit and return the final $\vec{v}^*$. \\

\textbf{Stopping Criteria}\\

Stopping criteria are used to speed up the convergence process. If our algorithm is converging to a solution and at each iteration the magnitude by which our solution is changing is tiny; then it is no longer beneficial for us to continue the loop and it may be better for us to break the loop. So we use two different criteria for the two different while loops. For the "outer" while loop we use the criteria:

\begin{align}
	\label{eq:Stopping Criteria Warp}
	&diff = \|I_1^{\omega} - I_0\|_2^2\nonumber\\
	&\frac{\|diff^{\omega} - diff^{\omega - 1}\|}{diff^\omega} < difference
\end{align}

which checks if the relative sum of the squared differences between the warped image and the target image between two iterations has only changed by a small amount called "\textit{difference}". If between two iterations it has only changed by a tiny margin then we assume that the loop has converged and we exit the loop. For this loop we set the parameter of "\textit{difference}" equal to $1\times10^{-3}$. For the "inner" while loop we use:

\begin{align}
	\label{eq:Stopping Criteria Iter}
	\frac{\|V_x^{k+1} - V_x^k\|_1}{\|V_x^k\|_1 + \epsilon} < tolerance, \frac{\|V_y^{k+1} - V_y^k\|_1}{\|V_y^k\|_1 + \epsilon} < tolerance
\end{align}

This loop has two stopping criteria and they must \textbf{both} be met in order to break the loop. These criteria check if the relative sum of absolute differences in the velocity components in both the $x$ and $y$ direction has only changed by a small margin we call "\textit{tolerance}". If it is less than "\textit{tolerance}" then we can exit the loop. In equation \ref{eq:Stopping Criteria Iter}, in the denominator we use a tiny almost insignificant value $\epsilon = 2^{-52}$ to avoid division by zero. The parameter for "\textit{tolerance}" is set equal to $1\times10^{-2}$. If these criteria above are met then a loop will be broken - this improves the efficiency of the algorithm and therefore decreases computation cost. \\

\textbf{Warping Function}\\

In Algorithm \ref{alg:Diffeomorphic Image Registration}, we say that we should get the warped source, $I_1^\omega$ by applying a $warp$ function, $warp(I_1, \vec{v}^\omega)$. To do this we get a deformation $\vec{v}^\omega$ from the algorithm and turn it into a displacement by using the identity grid $Id$. A displacement is then $Id + \vec{v}^\omega$. To warp, we then make use of Pytorch's grid sample function which which uses this displacement to sample the image at the corresponding displacement location. This can be clearly seen in figure \ref{fig:DiffIR Circle to C Deformation}. 

\section{Results}
\label{sec:Diffeomorphic Image Registration Results}

The algorithm that was used for diffeomorphic image registration was set out in Algorithm \ref{alg:Diffeomorphic Image Registration}, and just after that algorithm, we spoke about how it works and the different parameters that were used. We will now look at an example and see the output of this algorithm. The first example we will look at is a simple one so that we can really visualise what is being done. The source image for this example (the one which will be warped) is a white circle on a black background; while the target image will be a "c" shape in white on a black background. This example will not be very computationally expensive as there are a lot of matching black pixels which will not need to be warped at all. Furthermore, it will be easy to see what has happened - this will be compounded with the extra visualisation that we will later look at too. 

\begin{figure}[H]
	\begin{center}
		\includegraphics[width = 0.9\textwidth]{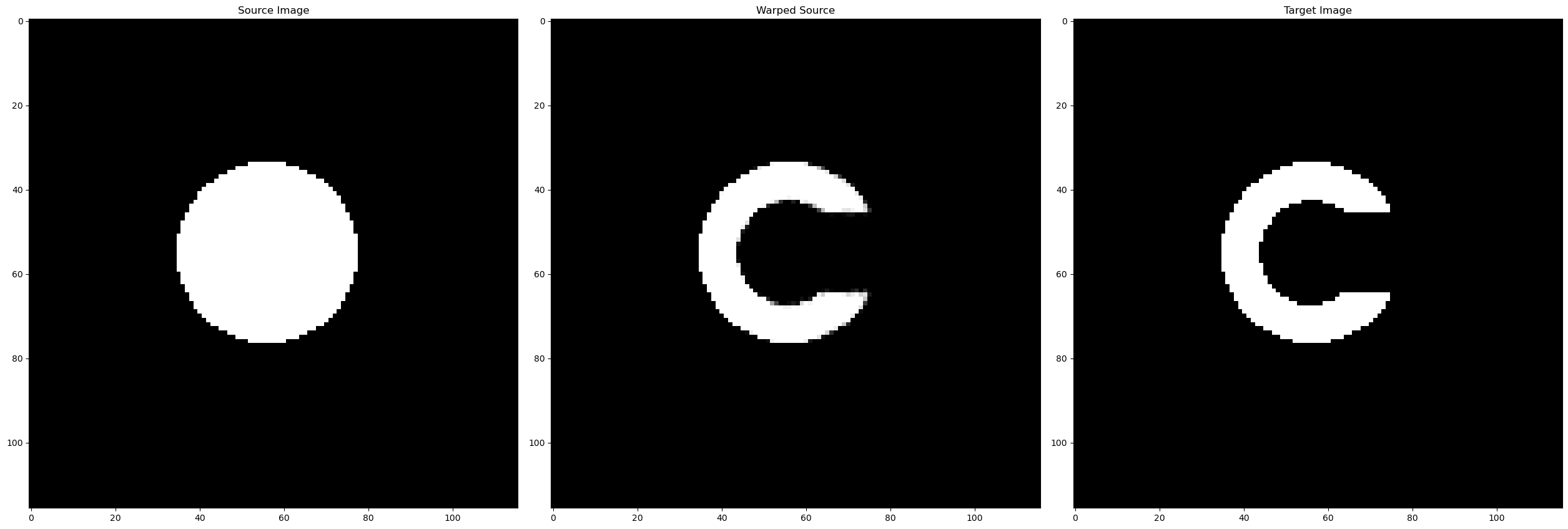}
		\caption{Diffeomorphic Image Registration of a Circle to a C} 
		\label{fig:DiffIR Circle to C}
	\end{center}
\end{figure}

In Figure \ref{fig:DiffIR Circle to C}, we see 3 distinct images all labelled. The one on the left is the source image (the circle), while the image on the far right is the target (what we want to warp the source image to look like). The image in the centre is what we labelled "Warped Source" and this is the output of our diffeomorphic image registration algorithm. As we see, it has deformed the source image to look exactly like the target image. While the algorithm \textit{can} output this image, we notice that the actual output of algorithm \ref{alg:Diffeomorphic Image Registration} is $\vec{v}$. This is a deformation field which describes how we move from the source image to the target image, and we can visualise this to see what exactly is going on. 

\begin{figure}[H]
	\centering
	\subfloat[Deformation Field to Warp a Circle to a C]{\includegraphics[width=0.49\textwidth]{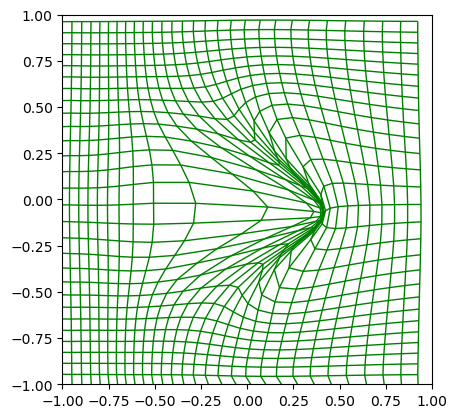}\label{fig:DiffIR Circle to C Deformation}}
	\subfloat[HSV of Deformation Field to Warp a Circle to a C]{\includegraphics[width=0.49\textwidth]{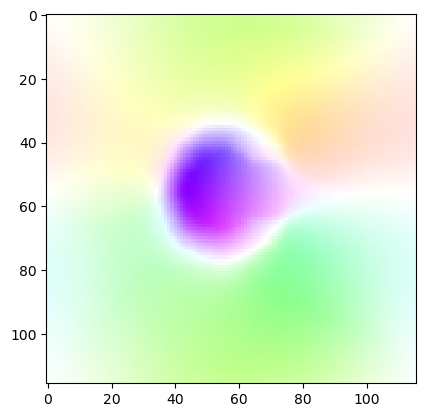}\label{fig:DiffIR Circle to C HSV}}
	\caption{Deformation Field for Circle to C Visualised}
	\label{fig:DiffIR Deformation Visualisation}
\end{figure}

In the above Figure \ref{fig:DiffIR Deformation Visualisation}, we see two figures which will help us visualise how we get from the source image in the above example, to the target image. On the left Figure \ref{fig:DiffIR Circle to C Deformation} we see the deformation field, while on the right Figure \ref{fig:DiffIR Circle to C HSV} we see what is called the HSV of that very deformation field. The deformation field can be thought of as showing you how the pixels are being moved, where a deformation field in a grid pattern would signify an identity transformation (doing nothing). The fact that this is not in a grid pattern shows exactly how the pixels are being transformed. The HSV, or Hue, Saturation, Value, is another way of visualising the deformation field. One way of thinking about it is that the HSV visualises the deformation is the polar plane using colours. For this, we use a constant saturation value for simplicity. Then imagine we have a colour wheel of varying colours, the Hue (angle) will decide which colour will be selected, and then the Value (magnitude) will decide the intensity of that colour. For instance, purple in figure \ref{fig:DiffIR Circle to C HSV} will represent going to the right or on a bearing of $90^\circ$, and this will mean that the deformation field at this position is pointing in that direction (this is the angle of that polar coordinate). The intensity of that colour will represent the magnitude of that polar coordinate. \\

We will now look at, in detail, an example on some cardiac MRI images, from the ACDC dataset \cite{dataset} and we will go into detail so that we can truly understand what is happening again, once we do this we will look at many examples to gain a complete summary of the results. The source image for this example will be MRI of a heart of a patient at End Diastole (ED), whereas the target image will be the MRI of that same heart at End Systole (ES) during the same cardiac cycle. 

\begin{figure}[H]
	\begin{center}
		\includegraphics[width = 0.9\textwidth]{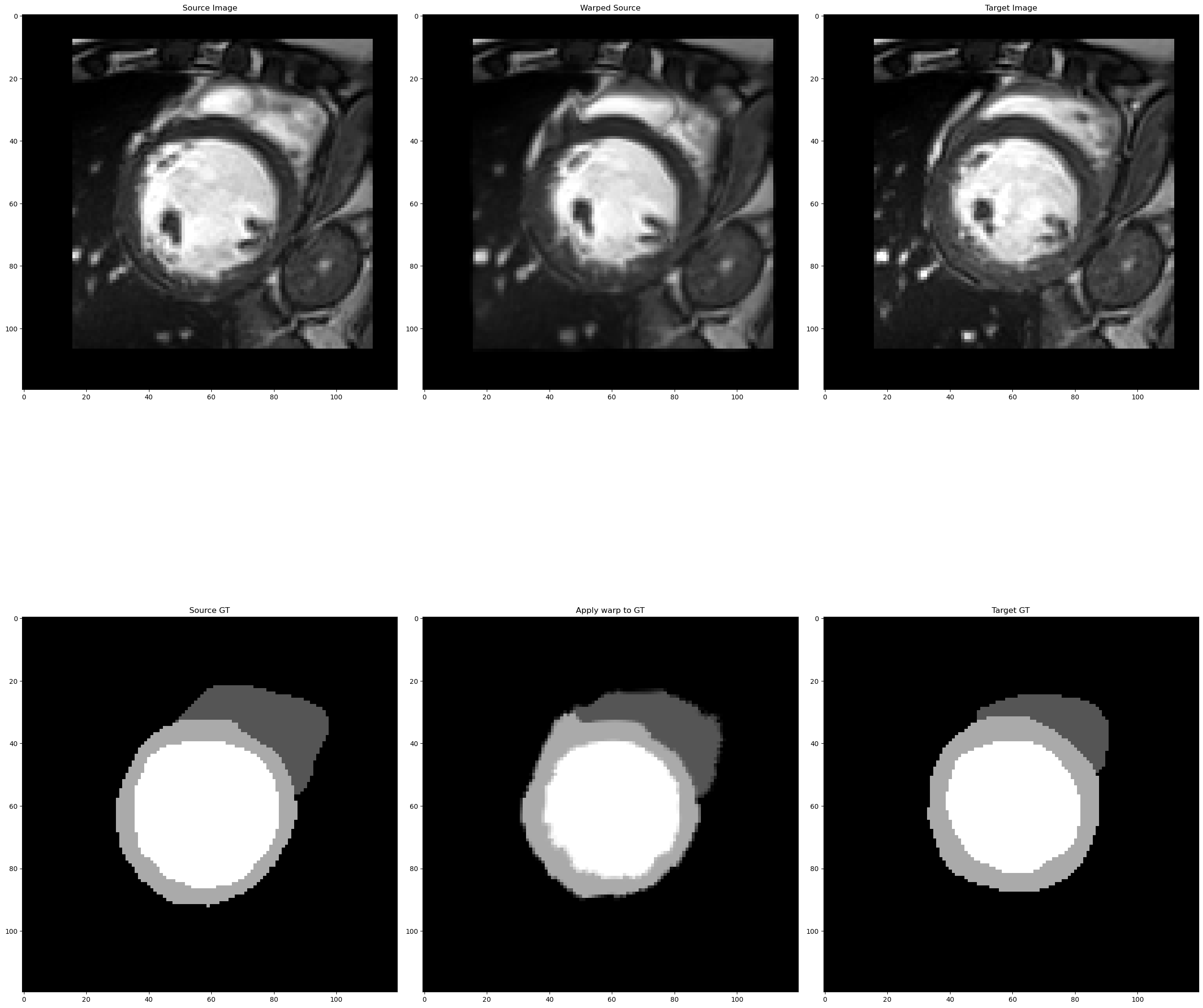}
		\caption{Diffeomorphic Image Registration on Cardiac MRI} 
		\label{fig:DiffIR Patient001}
	\end{center}
\end{figure}

For the first example we see in Figure \ref{fig:DiffIR Patient001} that there are $6$ images. The top left image is the source image of the heart in ED. The top right image is the target image of that same heart in ES. The top middle image is of the output of the diffeomorphic image registration algorithm using those source and target images. As we can see the top middle image resembles fairly closely the target image. The left ventricle (LV), the white circular part of the heart has decreased considerably in size from the source to the warped source images such that the shape resembles the target image. Furthermore, the right ventricle (RV), the long, white shape at the top of the image has changed shape too and looks similar to that in the target image. We do note that there is some inaccuracy particularly in the RV as there is a part of RV which looks "wavy". \\

The bottom images resemble the "ground truth" labels of the heart. In the bottom left image we see the ground truth labels for the source image that we used, where the white represents the left ventricle, the dark grey represents the right ventricle and the light grey represent the left ventricular myocardium. In the bottom right image we see the ground truth labels for the target image that we used. However, the bottom middle is not the output of Algorithm \ref{alg:Diffeomorphic Image Registration}, the bottom middle image actually resembles the deformation field that was outputted between the top left source and top right target images, when applied to the ground truth labelled image in the bottom left. So instead of applying the algorithm to the ground truth labels, we apply it to the unlabelled images, and then use that output to warp the ground truth labels. This is because in a real setting the ground truth labels may not always be accessible and warping on the ground truth labels gives much more accurate results. But it is interesting to see how each part of the heart is being warped by the Diffeomorphic Image Registration algorithm. For example, we can clearly see how the LV or RV is being warped by looking at the effect of the warp on the ground truth image. Furthermore, if we warp the ground truth source we can compare this to the ground truth target to empirically measure the accuracy of our algorithm using a similarity metric. 

\begin{figure}[H]
	\centering
	\subfloat[Deformation Field for DiffIR on Patient001]{\includegraphics[width=0.49\textwidth]{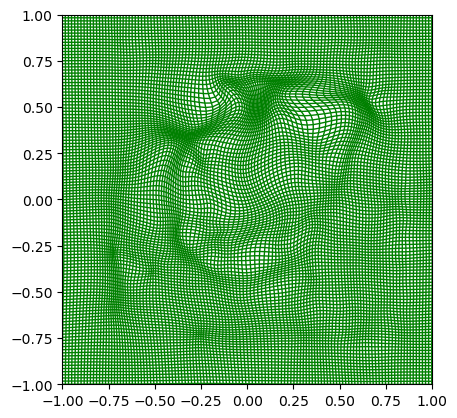}\label{fig:DiffIR Patient001 deformation}}
	\subfloat[HSV of Deformation Field for DiffIR on Patient001]{\includegraphics[width=0.49\textwidth]{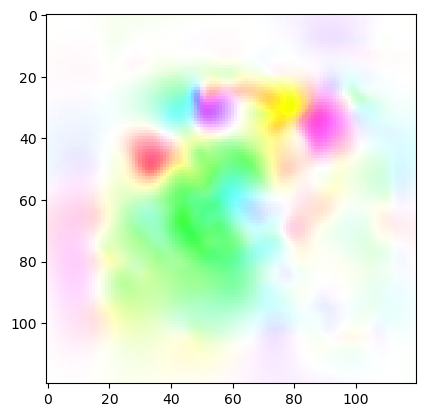}\label{fig:DiffIR Patient001 HSV}}
	\caption{Deformation Field for DiffIR on Patient001}
	\label{fig:DiffIR Patient001 Deformation Visualisation}
\end{figure}

Similarly, we again see the deformation field and the corresponding HSV plot for that field. These plots are to help us visualise what the algorithm is doing again, and this is important as we move onto complex examples warping cardiac images. 

\begin{figure}[H]
	\centering
	\subfloat{\includegraphics[width=0.98\textwidth]{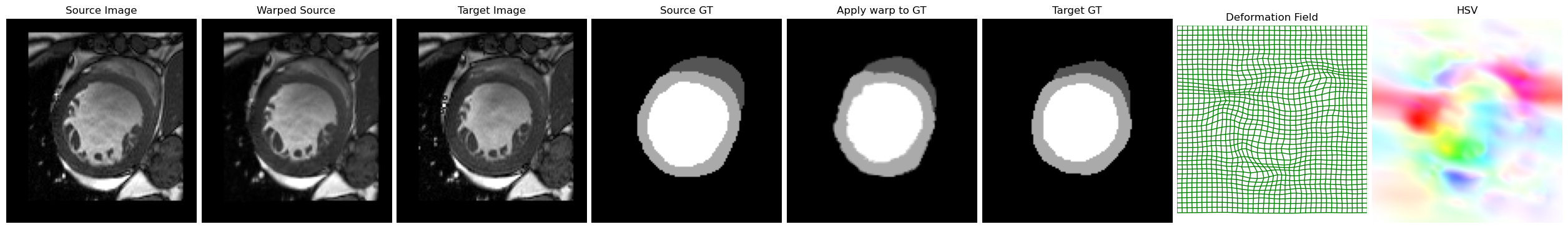}}
	\vspace*{-5mm}
	\subfloat{\includegraphics[width=0.98\textwidth]{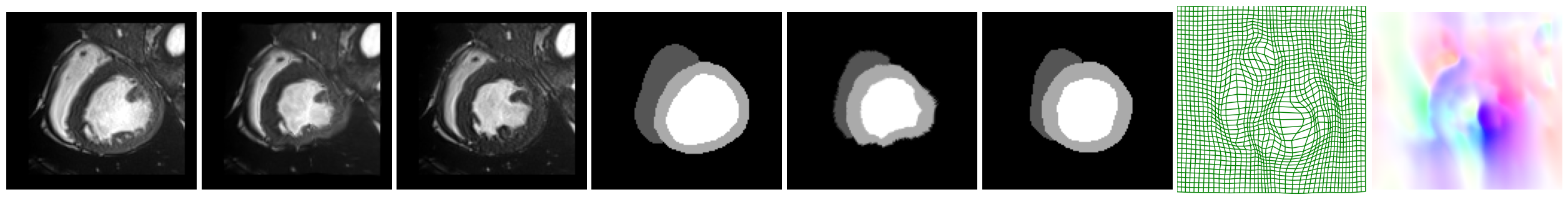}}
	\vspace*{-5mm}
	\subfloat{\includegraphics[width=0.98\textwidth]{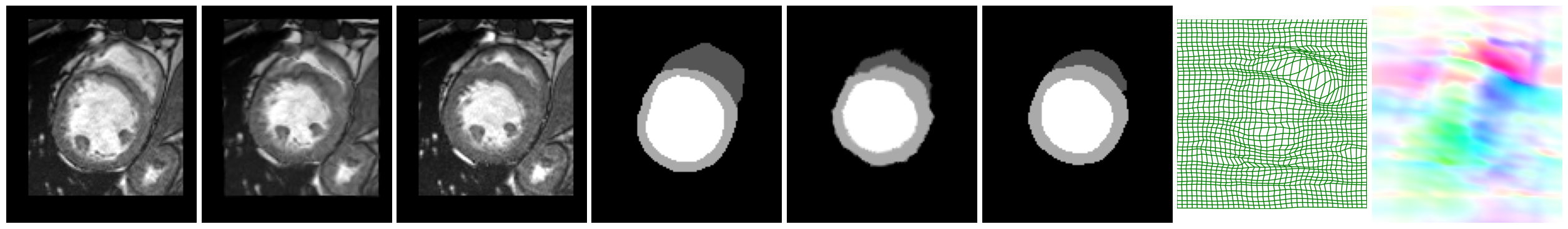}}
	\vspace*{-5mm}
	\subfloat{\includegraphics[width=0.98\textwidth]{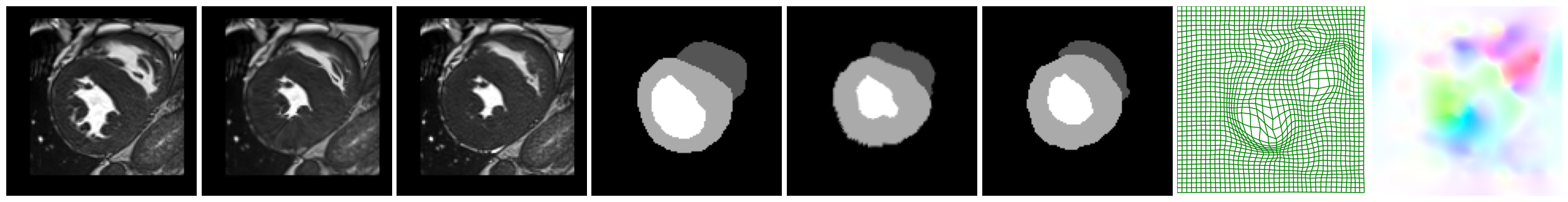}}
	\vspace*{-5mm}
	\subfloat{\includegraphics[width=0.98\textwidth]{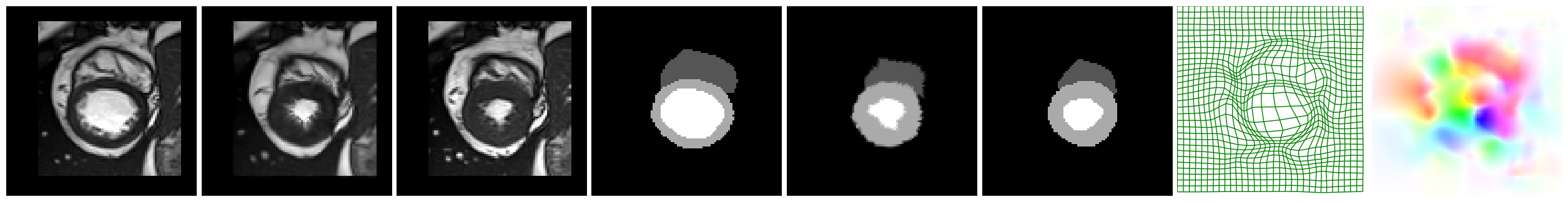}}
	\caption{Diffeomorphic Image Registration on Multiple Cardiac Patients}
	\label{fig:DiffIR Multiple Examples}
\end{figure}

In Figure \ref{fig:DiffIR Multiple Examples} above, we see multiple examples of our Diffeomorphic Image Registration algorithm with corresponding deformation fields and HSV images. While this is a good way to subjectively measure the success of our algorithm, it is not quantitative, and we need a way to objectively measure if our registration is good. To do this we use what is called the Dice coefficient, which is defined as:

\begin{align}
	\label{eq:Dice}
	DSC = \frac{2\|\vec{X}\cap\vec{Y}\|}{\|\vec{X}\| + \|\vec{Y}\|}
\end{align}

where $\|\vec{X}\|$ is the cardinality or number of elements in a set, or in our case image, $\vec{X}$. The Dice coefficient can be thought of as telling us the proportion of overlapping elements in one image when compared to another. This is where the deformation being applied to the source images ground truth labels is useful. As we can compare these images (column 5 in Figure \ref{fig:DiffIR Multiple Examples}), to the target images ground truth labels to calculate the Dice coefficient and objectively measure how accurate our registration is. The Dice coefficient returns a value in the range of 0 to 1, where 1 will imply a perfect match (and is returned if you test the dice score between and image and itself, since it overlaps itself perfectly) and 0 would imply no overlap between the two images. For the five examples in Figure \ref{fig:DiffIR Multiple Examples}, we get a Dice score of $0.8451$. While this is very good, you cannot only test it on 5 examples. For this reason we tested Algorithm \ref{alg:Diffeomorphic Image Registration} on all 100 images of the training section of the ACDC dataset \cite{dataset} and obtain a Dice score of $0.8170\pm0.072$. This score is significantly higher than the score posted in \cite{thorley2021nesterov} where they concluded in Table 1 that $\mathcal{L}^2 + 2^{nd}$O should have a mean average dice score of $0.751\pm0.045$. However, they also tested on $220$ examples not $100$ in my case, and the Cardiac MRI images were different too, our images are always sliced through the short-axis of the heart and both of these factors can lead to our perceived increase in Dice score.

\chapter{Motion Tracking}
\label{cha:Motion}

Before we move onto motion tracking, we need to ask ourselves, what is motion tracking, and how is it any different than diffeomorphic image registration? Diffeomorphic image registration was about aligning one image through warping to look like another. Motion tracking is concerned with by how much the warping is done and how it is warped through a sequence of images. In order to track motion accurately, you will want to know how something moves through time or through a sequence of images, and this will involve warping a sequence of images (or a group-wise warping) instead of a pairwise warping. Diffeomorphic image registration will not track motion as it will just tell us how to move something from A to B over a certain time frame, motion tracking will track exactly how and where A moved in order to get to B. For motion tracking, diffeomorphic image registration will be useful as we do still need to know how to warp two images together, however, we will make use of it in conjunction with what is called Robust PCA. 

\section{Robust PCA}
\label{sec:Robust PCA}

For motion tracking, we will make use of Robust Principle Component Analysis (RPCA) which was first proposed in \cite{candes2009robust}. Suppose we have a matrix, $\vec{M}$, and we wish to split that matrix into two other matrices; a low rank matrix $\vec{L}$, and a sparse matrix $\vec{S}$. \\

A low rank matrix, $\vec{L}$, is a matrix which has a \textit{low} rank where the rank of the matrix is defined as the number of linearly independent rows or columns of the matrix. So for a matrix to be low rank it must how a small number of linearly independent rows or columns. \\

A sparse matrix, $\vec{S}$, is defined as a matrix which has a lot of zero entries. So for a matrix to become \textit{more} sparse it must gain more zero entries. Or in the context of minimisation, the matrix must get less non-zero entries. \\

The motivation between wishing to split a matrix $\vec{M}$ into a low rank matrix, $\vec{L}$, and a sparse matrix, $\vec{S}$, such that $\vec{M} = \vec{L} + \vec{S}$, is as follows. Presume $\vec{M}$ is a noisy image, then hopefully $\vec{L}$ would be the image without noise, and $\vec{S}$ would be just the noise. This can be summarised in the following minimisation problem:

\begin{align}
	\label{eq:RPCA minimsation}
	\argmin_{\vec{L}, \vec{S}} \|\vec{L}\|_* + \lambda\|\vec{S}\|_1
\end{align}

Where $\|\vec{X}\|_*$ defines the nuclear norm which is defined as the sum of the singular values of $\vec{X}$ or in other words: $trace(\sqrt{\vec{A}^*\vec{A}})$ where $\vec{A}^*$ denotes the conjugate transpose of $\vec{A}$. Equation \ref{eq:RPCA minimsation} is exactly what we desire as if we make the rank of $\vec{L}$ smaller then the sum of it's singular values will decrease as if any columns or rows of $\vec{L}$ are linearly dependent then one of the singular values will be $0$. Furthermore, if $\vec{S}$ has less non-zero elements then the $L1$-norm will obviously decrease. Since we have the condition that $\vec{M} = \vec{L} + \vec{S}$ then this means that we can use the augmented Lagrange multipliers method again similarly to what we did in \ref{eq:Lagrangian HS}. 

\begin{align}
	\label{eq:Lagrangian RPCA}
	\argmin_{\vec{L}, \vec{S}} \|\vec{L}\|_* + \lambda\|\vec{S}\|_1 + \langle\vec{Y}, \vec{M} - \vec{L} - \vec{S}\rangle + \frac{\mu}{2}\|\vec{M} - \vec{L} - \vec{S}\|^2
\end{align}

where $\langle\vec{X}, \vec{Y}\rangle$ means to take the trace of $\vec{X}\vec{Y}$. Equation \ref{eq:Lagrangian RPCA} can again be solved using ADMM \cite{admm} where we solve for $\vec{L}$, then using that solution we solve for $\vec{S}$ and then finally update the Lagrange multiplier matrix $\vec{Y}$. This gives the following:

\begin{align}
	\label{eq:RPCA ADMM}
	\begin{cases}
		\vec{L}^{k+1} &= \argmin_\vec{L} \|\vec{L}^k\|_* + \langle\vec{Y}^k, \vec{M} - \vec{L}^k - \vec{S}^k\rangle + \frac{\mu}{2}\|\vec{M} - \vec{L}^k - \vec{S}^k\|^2\\
		\vec{S}^{k+1} &= \argmin_\vec{S} \lambda\|\vec{S}^k\|_1 + \langle\vec{Y}^k, \vec{M} - \vec{L}^{k+1} - \vec{S}^k\rangle + \frac{\mu}{2}\|\vec{M} - \vec{L}^{k+1} - \vec{S}^k\|^2\\
		\vec{Y}^{k+1} &= \vec{Y}^k + \mu(\vec{M} - \vec{L}^{k+1} - \vec{S}^{k+1})
	\end{cases}
\end{align}

To solve the following we need to make use of the shrinkage operator which is defined as: $\mathcal{S}_\tau(\vec{X}) = sign(x)\max(|x|-\tau, 0)$ where $x$ is an element within $\vec{X}$ and this operator is applied to every element within $\vec{X}$ individually so that we create another matrix of the same shape as $\vec{X}$. Furthermore, we define the singular value thresholding operator $\mathcal{D}_\tau(\vec{X}) = \vec{U}\mathcal{S}_\tau(\Sigma)\vec{V}^*$ where we apply the shrinkage operator to the singular values of the singular value decomposition of $\vec{X}$. \\

Now that we have defined the shrinkage operator and the singular value thresholding operator we can now solve equation \ref{eq:RPCA ADMM} using ADMM as follows:

\begin{numcases}{}
	\label{eq:RPCA L Solved}
	\vec{L}^{k+1} = \mathcal{D}_\mu(\vec{M} - \vec{S}^k - \mu^{-1}\vec{Y}^k)\\
	\label{eq:RPCA S Solved}
	\vec{S}^{k+1} = \mathcal{S}_{\lambda\mu}(\vec{M} - \vec{L}^{k+1} + \mu^{-1}\vec{Y})\\
	\label{eq:RPCA Y Solved}
	\vec{Y}^{k+1} = \vec{Y}^k + \mu(\vec{M} - \vec{L}^{k+1} - \vec{S}^{k+1})
\end{numcases}

These are the updates for how the Robust PCA algorithm works. Firstly, you define your image matrix $\vec{M}$ and initialise matrices $\vec{L}$, $\vec{S}$ and $\vec{Y}$ to be zero. Then you iterate through and update these solutions until you converge on a solution or until you reach the maximum number of iterations. 

\begin{algorithm}
	\caption{Robust PCA} \label{alg:Robust PCA}
	\textbf{Input Matrix:} $\vec{M}$\\
	\textbf{Input Parameters:} (maxiterations, $\lambda$, $\mu$, tolerance)\\
	\textbf{Initialise:} $\vec{L}^0 = \vec{0}, \vec{S}^0 = \vec{0}, \vec{Y}^0 = \vec{0}$\\
	\While{$k < \textnormal{maxiterations, or while Equation \ref{eq:Stopping Criteria RPCA} is not true}$}{
		Update $\vec{L}^{k+1}$ with Equation \ref{eq:RPCA L Solved}\\
		Update $\vec{S}^{k+1}$ with Equation \ref{eq:RPCA S Solved}\\
		Update $\vec{Y}^{k+1}$ with Equation \ref{eq:RPCA Y Solved}\\
	}
	\textbf{Return:} $\vec{L}, \vec{S}$
\end{algorithm}

For the Robust PCA Algorithm, \ref{alg:Robust PCA}, we need a stopping criteria again which will tell us if our algorithm is converging on a solution, if between iterations our solution is not changing much, then continuing the algorithm will waste computation power. The stopping criteria is defined as \cite{candes2009robust}:

\begin{align}
	\label{eq:Stopping Criteria RPCA}
	\frac{\|\vec{M} - \vec{L} - \vec{S}\|^2_F}{\|\vec{M}\|^2_F} \leq \textnormal{tolerance}
\end{align}

Now that we have the algorithm for Robust PCA, let's have a look at the output of the algorithm and explain what is happening and show that it works. \\

We tested this algorithm on a video of our own. The video consists of a still camera (this is important) and me running through a field. This video is vectorised and then reshaped back into a $2$-D matrix (more details on this in Section \ref{sec:Beyond Pairwise}). This we then set to $\vec{M}$, the input to the Robust PCA Algorithm \ref{alg:Robust PCA}. We set the "maxiterations" at $300$ and a tolerance of $1e-7$. $\lambda$ is defined to be $\frac{1}{3max(shape(\vec{M}))}$ where $shape$ will have two outputs $m,n$ that correspond to the size of the dimensions of $\vec{M}$. We then set $\mu = 10\lambda$. The video consisted of $48$ frames of $540 \times 960$ images. This was the output from the Robust PCA algorithm:

\begin{figure}[H]
	\centering
	\subfloat[Source Frame 0]{\includegraphics[width=0.33\textwidth]{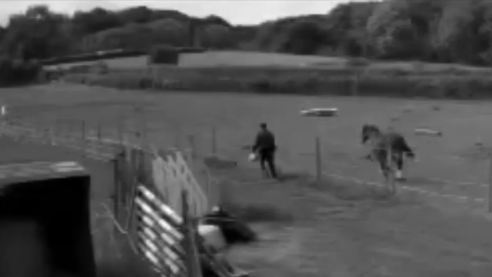}\label{fig:rpca source 0}}
	\subfloat[Low Rank Frame 0]{\includegraphics[width=0.33\textwidth]{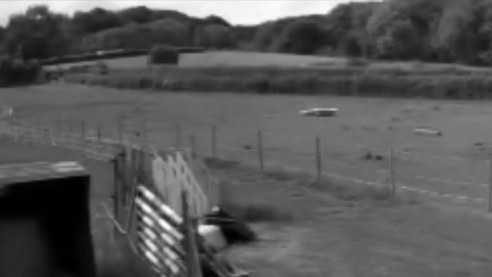}\label{fig:rpca low rank 0}}
	\subfloat[Sparse Frame 0]{\includegraphics[width=0.33\textwidth]{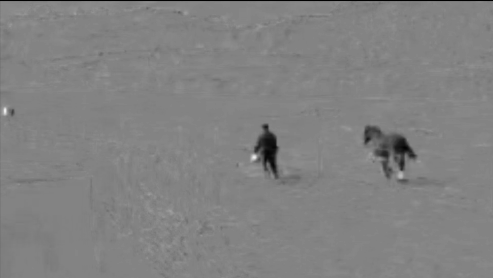}\label{fig:rpca sparse 0}}
	\hfill
	\subfloat[Source Frame 22]{\includegraphics[width=0.33\textwidth]{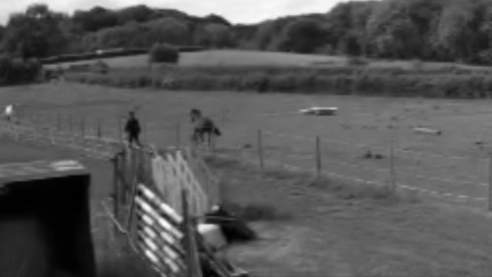}\label{fig:rpca source 22}}
	\subfloat[Low Rank Frame 22]{\includegraphics[width=0.33\textwidth]{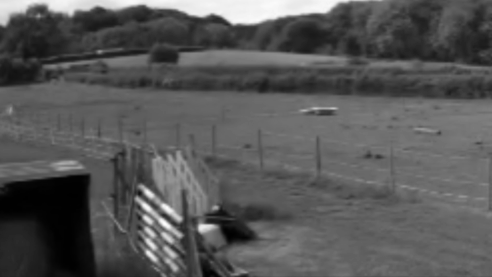}\label{fig:rpca low rank 22}}
	\subfloat[Sparse Frame 22]{\includegraphics[width=0.33\textwidth]{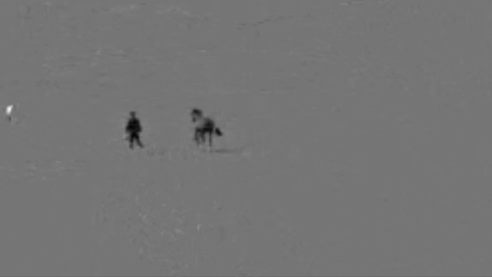}\label{fig:rpca sparse 22}}
	\hfill
	\subfloat[Source Frame 47]{\includegraphics[width=0.33\textwidth]{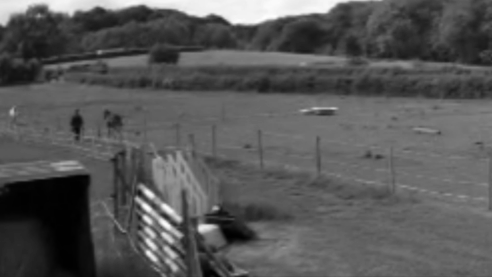}\label{fig:rpca source 47}}
	\subfloat[Low Rank Frame 47]{\includegraphics[width=0.33\textwidth]{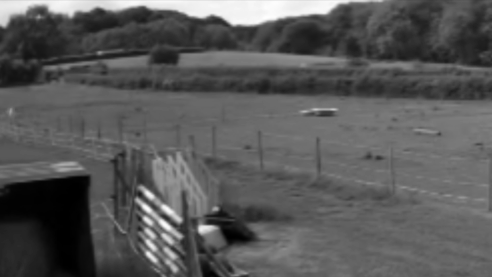}\label{fig:rpca low rank 47}}
	\subfloat[Sparse Frame 47]{\includegraphics[width=0.33\textwidth]{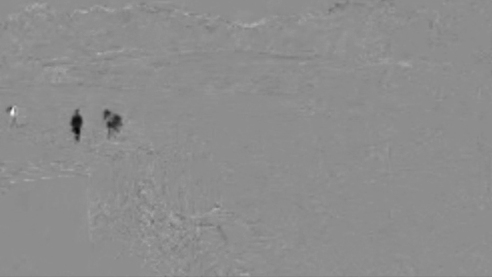}\label{fig:rpca sparse 47}}
	\caption{Robust PCA output}
	\label{fig:Robust PCA Application}
\end{figure}

In the left hand column of Figure \ref{fig:Robust PCA Application}, we see the original source images, which contain the background and me running on the scene too. On the middle column we see the low rank decomposition of these frames - this contains only the background. This makes it low rank as these frames look identical as they only contain the background and nothing else. In the right hand column we see the sparse decomposition. The sparse decomposition is clearly just the "stuff" within the image that is moving and such can be separated from the background. These sparse decompositions contain just me (and some noise). While some noise may be the result of terminating the algorithm early, it can also be the result of the fact that the initial video was just noisy. To reduce the effect of this we can perform some sort of blurring or noise reduction to the initial frames. 

\section{Beyond Pairwise} 
\label{sec:Beyond Pairwise}

Up until now we have only discussed up to the prospect of pairwise operations, in the case of diffeomorphic image registration this means warping one image to look like another - and with Robust PCA the method seems fairly vague. To elaborate more about Robust PCA, and to move on from pairwise image registration, we must discuss group-wise operations. To begin, we will explain what this means in the context of image registrations, then we will explain it in the context of Robust PCA to solidify our understanding the Robust PCA algorithm. Finally, we will explain how we move on from this and how this helps our task of motion tracking.\\ 

\textbf{Image Registration}\\

Currently, we have only discussed pairwise image registration - this entails warping a source image onto a target image. We have showed that this works and while this is excellent, it may not necessarily be exactly what we want. If we were to use cardiac data, for instance, it will usually come in the form of a video. Instead of warping one image from the end systole to another image at the end of diastole, it would be more useful to warp ALL of the images throughout the entire cardiac cycle to a reference image at the end of diastole, say. As a video can be thought of as a sequence of images, this would mean that we would have a sequence of source images $\{I_1, I_2, ... I_t\}$ that we would like to warp to a target image $I_0$. Theoretically you could just register the images one after another but this would be tedious. Instead, we could combine the images in a intuitive way. What we do is transform each of our images with shape $m \times n$ into vectors of shape $mn \times 1$ and if we applied this to say $I_1$, the vectorised form would be denoted as: $\vectorise(I_1)$. If we vectorise each image, we can then concatenate them such that we create a 2D matrix, $\vec{M}$, with shape $mn \times t$, where $t$ is the number of source images that we have, such that: $$\vec{M} = [\vectorise(I_1), \vectorise(I_2), \dots , \vectorise(I_t)]$$
This new matrix can be considered our new source "image" and then we just have to create a new target "image". To do this we do a similar procedure. First we vectorise the target image to a shape of $nm \times 1$, but since all the source images will have the same target image, we just need to repeat this vector $t$ times so that we have a target "image" $[\vectorise(I_0), \vectorise(I_0), \dots , \vectorise(I_0)]$ with shape $nm \times t$. Now that we have these new source and target "images" we can just continue with image registration as usual. This technique of making use of vectorisation is very important and we will make use of this to explain how the Robust PCA algorithm works. \\

\textbf{Robust PCA}\\

During our discussion of Robust PCA in Section \ref{sec:Robust PCA} we spoke about the process in very general terms. We know that the aim of Robust PCA is to split a matrix $\vec{M}$ into a low rank matrix and a sparse matrix, $\vec{L}$ and $\vec{S}$ respectively. We know what it means for a matrix to be low rank. But in this context what does it mean exactly? If $\vec{M}$ is just a normal image, it could be a heart or an animal or anything, what exactly does a low rank version of that image look like? How do you decrease the rank of that image? If the image were of a pattern with noise then of course you could decrease the rank by removing the noise, the rank will then be low because the image has a pattern. However, if your image is of something that isn't a pattern, then these questions themselves don't make sense - and this is because we wouldn't \textit{want} a low rank version of this image as it wouldn't mean much to us. To explain why a low rank version of any image isn't particularly useful, I will explain the proper way that we use to get a low rank matrix, and then contextualise this in comparison to the above example which doesn't make much sense. \\

Suppose we have a video from a stationary camera of someone running through a field. Like we spoke about for group-wise image registration, we can vectorise each frame of this image to create a new matrix where each column represents a frame of the video. Now we can ask that question again, what does a low rank version of \textit{this} matrix look like? Now, this question has an answer. If we wanted to decrease the rank of this matrix, we need each column (a frame of the video) to be the same. If the video is of a person running through a field, then the part of the video which doesn't change is the background (not the person running). So a low rank version of this matrix will be just the background. Since we have the condition that $\vec{M} = \vec{L} + \vec{S}$ we know that the sparse matrix must in this example contain just the person running \cite{lowranksparsecardiac}.\\

Now we can see the difference between these two examples, in the second example we can clearly see that a low rank version of a \textit{video} will be just the background, whereas a low rank version of an \textit{image} doesn't really make sense? You can't say it will be the background again as that might not be what decreases the rank. It is important to note that this doesn't necessarily mean that you cannot decrease the rank of just an image; you could do that. It is just important to state that the context matters, and that in the context of this paper, a low rank image of say a heart, doesn't mean much as what is it low rank in respect to? Why is it low rank?\\

For motion tracking we will make use of both of these concepts as we will want to perform groupwise image registration. 

\section{Motion Tracking} 
\label{sec:Motion Tracking} 

In Section \ref{sec:Robust PCA}, we discussed the framework of how Robust PCA works and the desire to split a matrix into a low rank matrix and a sparse matrix. This idea can be extended to help us do motion tracking by doing Robust PCA in conjunction with diffeomorphic image registration. The alignment of one sequence of images to another sequence is defined to be good if the rank of the output after alignment is low \cite{haase2020deformable} - this is a perfect place to use Robust PCA.\\

From Section \ref{sec:Beyond Pairwise}, we know that we will want to perform group-wise operations instead of pairwise, and we will make use of the methods that were laid out in that section. To begin with we will first define our source images matrix $\vec{M}_1 = [\vectorise(I_1), \vectorise(I_2), \dots , \vectorise(I_t)]$ such that $\vec{M}_1$ has all source images vectorised and concatenated together. We also need to register these images to a target matrix, so we will define another matrix $\vec{M}_0$ such that $\vec{M}_0 = [\vectorise(I_0), \vectorise(I_0), \dots , \vectorise(I_0)$ where this matrix has the target image vectorised and copied $t$ times so that the matrix also has shape $mn \times t$. However, we need $\vec{M}_1$ to be our warped source images and so each of these vectorised images depend on our velocity $\vec{v}^k_t$ at iteration $k$ of our registration for the $t$-th image. It would be more accurate to define $\vec{M}_1$ as the following:

\begin{align}
	\label{eq:Casorati Matrix M1}
	\vec{M}_1^k = [\vectorise(I_1(\vec{x} + \vec{v}^k_1)), \vectorise(I_2(\vec{x} + \vec{v}^k_2)), \dots , \vectorise(I_t(\vec{x} + \vec{v}^k_t))]
\end{align}

We then wish to define the difference of $\vec{M}_1$ and $\vec{M}_0$ as this will make a new matrix $\vec{M}$ that is similar in nature to $\rho(\vec{v})$ that appeared in Equation \ref{eq:Variational HS}:

\begin{align}
	\label{eq:Casorati Matrix}
	\vec{M}^k = \vec{M}_1^k - \vec{M}_0
\end{align}

We want to make use of Robust PCA and we also know that a good registration will result in an output which is low rank - this means we will decompose $\vec{M}$ into a low rank matrix, $\vec{L}$, and a sparse matrix, $\vec{S}$. This can also be thought of as decomposing $\rho(\vec{v})$ from Equation \ref{eq:Variational HS} into a low rank and sparse matrix (obviously in this case we are dealing with source and target matrices). Furthermore, to promote smooth deformations we will make use of a regularisation term, the one we used before in Equation \ref{eq:Variational HS} will be sufficient. Finally, we will decouple $\vec{v}$ using parameter $\vec{w}$ as we did in Equation \ref{eq:Decoupled Variational HS} such that $\vec{w} = \vec{v}$. This will make our new optimisation problem take the form:

\begin{align}
	\label{eq:Motion Tracking Minimisation}
	\argmin_\vec{v} \|\vec{L}\|_* + \mu\|\vec{S}\|_1 + \frac{\lambda}{2}\|\nabla^n \vec{w}\|^2 : \vec{M} = \vec{L} + \vec{S}, \vec{w} = \vec{v}
\end{align}

This is a constrained optimisation problem so we can make use of our augmented Lagrangian parameters like we did in Equation \ref{eq:Lagrangian HS}. Furthermore, since our problem is very similar to the Robust PCA problem we can expand it similarly with augmented Lagrange parameters as we did in Equation \ref{eq:Lagrangian RPCA}:

\begin{align}
	\label{eq:Motion Tracking Lagrangian}
	\mathcal{L}_\mathcal{A}= \|\vec{L}\|_* + \mu\|\vec{S}\|_1 + \langle\vec{Y}, \vec{M} - \vec{L} - \vec{S}\rangle + \frac{\rho}{2}\|\vec{M} - \vec{L} - \vec{S}\|^2 + \frac{\lambda}{2}\|\nabla^n\vec{w}\|^2 + \frac{\theta}{2}\|\vec{w} - \vec{v} - \vec{b}\|^2
\end{align}

where $\mu, \rho, \lambda, \theta > 0$. As we saw in section \ref{sec:Robust PCA} the operator $\langle\vec{X},\vec{Y} \rangle$ means to take the trace of $\vec{X}\vec{Y}$. This term makes solving this problem quite difficult as so we can complete the square of equation \ref{eq:Motion Tracking Lagrangian} to combine this difficult term with the term following it giving us:

\begin{align}
	\label{eq:Motion Tracking Lagrangian Complete Square}
	\mathcal{L}_\mathcal{A}= \|\vec{L}\|_* + \mu\|\vec{S}\|_1 + \frac{\rho}{2}\|\vec{M} - \vec{L} - \vec{S} + \frac{1}{\rho}\vec{Y}\|^2 + \frac{\lambda}{2}\|\nabla^n\vec{w}\|^2 + \frac{\theta}{2}\|\vec{w} - \vec{v} - \vec{b}\|^2
\end{align}

By making using of ADMM \cite{admm} we can separate the above Equation \ref{eq:Motion Tracking Lagrangian Complete Square} into $6$ sub-problems, $4$ of which will need to be solved for, while the other $2$ are our Lagrangian parameters which have a set solution, these are $\vec{Y}$ and $\vec{b}$. We separate the sub-problems out by looking at the respective terms that depend on each of the variables, for example the sub-problem for $\vec{L}$ will be the terms of Equation \ref{eq:Motion Tracking Lagrangian Complete Square} which have $\vec{L}$ appearing in them. This gives the following set of Equations:

\begin{align}
	\label{eq:Motion Tracking Lagrangian ADMM Cases}
	\begin{cases}
		\vec{L}^{k+1} &= \underset{\vec{L}}{\argmin}\hspace{5pt} \|\vec{L}^k\|_* + \frac{\rho}{2}\|\vec{M}^k - \vec{L}^k - \vec{S}^k + \frac{1}{\rho}\vec{Y}^k\|^2 \\
		\vec{S}^{k+1} &= \underset{\vec{S}}{\argmin}\hspace{5pt} \mu\|\vec{S}\|_1 + \frac{\rho}{2}\|\vec{M}^k - \vec{L}^k - \vec{S}^k + \frac{1}{\rho}\vec{Y}^k\|^2 \\
		\vec{v}^{k+1} &= \underset{\vec{v}}{\argmin}\hspace{5pt} \frac{\rho}{2}\|\vec{M}^k - \vec{L}^k - \vec{S}^k + \frac{1}{\rho}\vec{Y}^k\|^2 + \frac{\theta}{2}\|\vec{w}^k - \vec{v}^k - \vec{b}^k\|^2 \\
		\vec{w}^{k+1} &= \underset{\vec{w}}{\argmin}\hspace{5pt} \frac{\lambda}{2}\|\nabla^n\vec{w}^k\|^2  + \frac{\theta}{2}\|\vec{w}^k - \vec{v}^k - \vec{b}^k\|^2 \\
		\vec{Y}^{k+1} &= \vec{Y}^k + \rho(\vec{M}^k - \vec{L}^k - \vec{S}^k) \\
		\vec{b}^{k+1} &= \vec{b}^k + \vec{v}^k - \vec{w}^k
	\end{cases}
\end{align}

As stated previously, the bottom two equations of Equation \ref{eq:Motion Tracking Lagrangian ADMM Cases} are deterministic and we know these are correct from \cite{candes2009robust}, \cite{admm}. The $4$ equations above this need to be solved and we need to calculate a closed form solution for each of them. For some of these the solution is fairly easy and for others it is a bit more complex - we shall go through them one at a time, however, we will deal with the $\vec{v}$ sub-problem last as it is the most difficult and requires multiple steps. The solutions for $\vec{L}$ and $\vec{S}$ are easily solved as they are very closely related to how they are solved for in Robust PCA \cite{candes2009robust}, \cite{cai2008singular}. These closed form solutions are clearly:

\begin{numcases}{}
	\label{eq:Motion Tracking L Solved}
	\vec{L}^{k+1} = \mathcal{D}_{\frac{1}{\rho}}(\vec{M}^k - \vec{S}^k - \frac{1}{\rho}\vec{Y}^k)\\
	\label{eq:Motion Tracking S Solved}
	\vec{S}^{k+1} = \mathcal{S}_{\frac{\mu}{\rho}}(\vec{M}^k - \vec{L}^{k+1} + \frac{1}{\rho}\vec{Y}^k)
\end{numcases}

These are our closed form solutions for $\vec{L}$ and $\vec{S}$. Similarly to in Section \ref{sec:Robust PCA}, $\mathcal{D}$ denotes the singular value thresholding operator and $\mathcal{S}$ denotes the shrinkage operator (please refer to Section \ref{sec:Robust PCA} for more details). Now we can move onto our next sub-problems. We will tackle the $\vec{w}$ sub-problem next. While it is complicated to solve, we realise that we do not need to solve it as we have already done this before. If we look at the $\vec{w}$ sub-problem in Equation \ref{eq:Motion Tracking Lagrangian ADMM Cases}, we notice that it is identical to the problem in Equation \ref{eq:Lagrangian Expanded}. This means that the closed form solution will be the same - the method to solving this was set out in Section \ref{sec:Diffeomorphic Image Registration}:

\begin{align}
	\label{eq:Motion Tracking w Solved}
	\vec{w}^{k+1}  = \mathcal{F}^{-1}\left(\frac{\theta\mathcal{F}(\hat{\vec{v}}^k+\vec{b}^k)}{\lambda\mathcal{F}(\Delta^n) + \theta}\right)
\end{align}

Again, this means we have the exact same notation, meaning that $\mathcal{F}$ denotes the discrete Fourier transform which has been chosen due to our periodic boundary conditions, and $\Delta^n$ denotes the $n$-th order Laplace operator (more details on these were set out in Section \ref{sec:Diffeomorphic Image Registration}). We will now tackle the $\vec{v}$ sub-problem of Equation \ref{eq:Motion Tracking Lagrangian ADMM Cases}. To begin with, let us remind ourselves of the optimisation problem:

$$\vec{v}^{k+1} = \underset{\vec{v}}{\argmin}\hspace{5pt} \frac{\rho}{2}\|\vec{M}^k - \vec{L}^k - \vec{S}^k + \frac{1}{\rho}\vec{Y}^k\|^2 + \frac{\theta}{2}\|\vec{w}^k - \vec{v}^k - \vec{b}^k\|^2$$

This problem is fairly complex and to solve it we will consider a slightly simpler problem and then extend our solution quite easily to our problem. First, instead of $\vec{M}$ denoting a group of images; instead just pretend, for now, that we are dealing with pairwise images, $I_1(\vec{x} + \vec{v}^k)$, and $I_0(\vec{x})$, say. Now we notice we have another problem to deal with, $I_1(\vec{x} + \vec{v}^k)$, is non-linear. To solve this problem we must linearise this term using a Taylor approximation. This gives us:

$$\vec{M}^{k+1} \approx I_1(\vec{x} + \vec{v}^k) + \vec{J}\vec{v}^k - I_0(\vec{x})$$

where $I_0(\vec{x})$ refers to our source image. Furthermore, we have that $\vec{J}$ is the derivative of $I_1(\vec{x} + \vec{v}^k)$, this will be a vector as we will have the derivative in both the $x$ and $y$ directions. Let us substitute this approximation into our sub-problem to get the following linearised sub-problem. 

$$\vec{v}^{k+1} = \underset{\vec{v}}{\argmin}\hspace{5pt} \frac{\rho}{2}\|I_1(\vec{x} + \vec{v}^k) + \vec{J}\vec{v}^k - I_0(\vec{x}) - \vec{L}^k - \vec{S}^k + \frac{1}{\rho}\vec{Y}^k\|^2 + \frac{\theta}{2}\|\vec{w}^k - \vec{v}^k - \vec{b}^k\|^2$$

Now this problem can be solved simply, if we differentiate and set equal to $0$ and solve for $\vec{v}$ then we will have a closed-form solution for $\vec{v}$:

\begin{align}
	\vec{J}\rho(I_1(\vec{x} + \vec{v}) + \vec{J}\vec{v} - I_0(\vec{x}) - \vec{L} - \vec{S} + \frac{1}{\rho}\vec{Y}) + \theta(\vec{v} + \vec{b} - \vec{w}) = 0 \nonumber \\
	\rho\vec{J}\vec{J}^T\vec{v} + \theta\vec{v} + \vec{J}\rho(I_1(\vec{x} + \vec{v}) - I_0(\vec{x}) - \vec{L} - \vec{S} + \frac{1}{\rho}\vec{Y}) + \theta(\vec{b} - \vec{w}) = 0 \nonumber \\
	[\rho\vec{J}\vec{J}^T + \theta\mathbb{1}]\vec{v} = \theta(\vec{w} - \vec{b}) + \rho\vec{J}(\vec{L} + \vec{S} - \frac{1}{\rho}\vec{Y}) - \rho\vec{J}(I_1(\vec{x} + \vec{v}) - I_0(\vec{x})) \nonumber
\end{align}

This is not quite yet the closed form solution for $\vec{v}$. To arrive at this we need to calculate the inverse of $[\rho\vec{J}\vec{J}^T + \theta\mathbb{1}]$. As we did in Section \ref{sec:Diffeomorphic Image Registration} we will do this by using the Sherman-Morrison equation to calculate the inverse. 

\begin{align}
	(A + uv^T)^{-1} &= A^{-1} - \frac{A^{-1}uv^TA^{-1}}{1 + v^TA^{-1}u} \nonumber \\
	(\theta \mathbb{1} + \rho\vec{J}\vec{J}^T)^{-1} &= (\theta \mathbb{1} )^{-1} - \frac{(\theta \mathbb{1} )^{-1}\rho\vec{J}\vec{J}^T(\theta \mathbb{1} )^{-1}}{1 + \rho\vec{J}^T(\theta \mathbb{1} )^{-1}\vec{J}}\nonumber\\
	&= \frac{\mathbb{1}}{\theta} - \frac{\frac{\mathbb{1}}{\theta}\rho\vec{J}\vec{J}^T\frac{\mathbb{1}}{\theta}}{1 + \rho\vec{J}^T\frac{\mathbb{1}}{\theta}\vec{J}}\nonumber\\
	&= \frac{\mathbb{1}}{\theta} - \frac{\frac{\rho\vec{J}\vec{J}^T}{\theta^2}}{1 + \frac{\rho\vec{J}^T\vec{J}}{\theta}}\nonumber\\
	&= \frac{\mathbb{1}}{\theta} - \frac{\frac{\rho\vec{J}\vec{J}^T}{\theta^2}}{\frac{\theta + \rho\vec{J}^T\vec{J}}{\theta}}\nonumber\\
	&= \frac{\mathbb{1}}{\theta} - \frac{\rho\vec{J}\vec{J}^T}{\theta(\theta + \rho\vec{J}^T\vec{J})}\nonumber\\
	&= \frac{(\mathbb{1}\theta + \mathbb{1}\rho\vec{J}^T\vec{J}) - \rho\vec{J}\vec{J}^T}{\theta(\theta + \rho\vec{J}^T\vec{J})}\nonumber\\
	&= \frac{\begin{bmatrix}
			\rho\biggl(\frac{\partial I}{\partial x}\biggr)^2 + \rho\biggl(\frac{\partial I}{\partial y}\biggr)^2 + \theta & 0 \\
			0 & \rho\biggl(\frac{\partial I}{\partial x}\biggr)^2 + \rho\biggl(\frac{\partial I}{\partial y}\biggr)^2 + \theta
		\end{bmatrix} - \rho\begin{bmatrix}
			\biggl(\frac{\partial I}{\partial x}\biggr)^2 & \frac{\partial I}{\partial x}\frac{\partial I}{\partial y} \\
			\frac{\partial I}{\partial y}\frac{\partial I}{\partial x} & \biggl(\frac{\partial I}{\partial y}\biggr)^2
	\end{bmatrix}}{\theta\biggl(\rho\biggl(\frac{\partial I}{\partial x}\biggr)^2 + \rho\biggl(\frac{\partial I}{\partial y}\biggr)^2 + \theta\biggr)}\nonumber\\
	\label{eq:Motion Tracking Sherman-Morrison Inverse}
	&=\frac{\begin{bmatrix}
			\biggl(\rho\biggl(\frac{\partial I}{\partial y}\biggr)^2 + \theta\biggr) & -\rho\biggl(\frac{\partial I}{\partial x}\frac{\partial I}{\partial y}\biggr) \\
			-\rho\biggl(\frac{\partial I}{\partial y}\frac{\partial I}{\partial x}\biggr) & \biggl(\rho\biggl(\frac{\partial I}{\partial x}\biggr)^2 + \theta\biggr)
	\end{bmatrix}}{\theta\biggl(\rho\biggl(\frac{\partial I}{\partial x}\biggr)^2 + \rho\biggl(\frac{\partial I}{\partial y}\biggr)^2 + \theta\biggr)}
\end{align}

Equation \ref{eq:Motion Tracking Sherman-Morrison Inverse} is the solution for $(\theta \mathbb{1} + \rho\vec{J}\vec{J}^T)^{-1}$ and we can use this to get the closed form solution for $\vec{v}$. We said that we were dealing with a pretend simple example in which we replaced $\vec{M}$ with a pair of images and then with $I_1(\vec{x} + \vec{v}^k) + \vec{J}\vec{v}^k - I_0(\vec{x})$. Using this, we can see that this example that we took on clearly extends to the harder problem as instead of having $I_1$ and $I_0$, we have $\vec{M}_1$ and $\vec{M}_0$, and the difference of these terms is precisely $\vec{M}$. This means we can display all the closed form solutions, as well as the updates for $\vec{b}$ and $\vec{Y}$ in the following:

\begin{numcases}{}
	\label{eq:Motion Tracking L cases}
	\vec{L}^{k+1} = \mathcal{D}_{\frac{1}{\rho}}(\vec{M}^k - \vec{S}^k - \frac{1}{\rho}\vec{Y}^k)\\
	\label{eq:Motion Tracking S cases}
	\vec{S}^{k+1} = \mathcal{S}_{\frac{\mu}{\rho}}(\vec{M}^k - \vec{L}^{k+1} + \frac{1}{\rho}\vec{Y}^k)\\
	\label{eq:Motion Tracking v cases}
	\vec{v}^{k+1} = (\theta \mathbb{1} + \rho\vec{J}\vec{J}^T)^{-1}\biggl(\theta(\vec{w}^k - \vec{b}^k) + \rho\vec{J}(\vec{L}^{k+1} + \vec{S}^{k+1} - \frac{1}{\rho}\vec{Y}^k) - \rho\vec{J}(\vec{M}^k_1 - \vec{M}_0)\biggr)\\
	\label{eq:Motion Tracking v hat cases}
	\hat{\vec{v}}^{k+1} = \alpha\vec{v}^{k+1} + (1-\alpha)\vec{w}^{k}\\
	\label{eq:Motion Tracking w cases}
	\vec{w}^{k+1}  = \mathcal{F}^{-1}\left(\frac{\theta\mathcal{F}(\hat{\vec{v}}^{k+1}+\vec{b}^k)}{\lambda\mathcal{F}(\Delta^n) + \theta}\right)\\
	\label{eq:Motion Tracking Y cases}
	\vec{Y}^{k+1} = \vec{Y}^k + \rho(\vec{M}^k - \vec{L}^{k+1} - \vec{S}^{k+1}) \\
	\label{eq:Motion Tracking b cases}
	\vec{b}^{k+1} = \vec{b}^k + \hat{\vec{v}}^{k+1} - \vec{w}^{k+1}
\end{numcases}

We make use of the over-realxed ADMM method again like what was done in \cite{arbitraryordertotalvariation}, this is why we have an extra update for $\hat{\vec{v}}^{k+1}$ that was not previously mentioned but now appears in the closed form solutions. With these closed form solutions we can now set out the algorithm used for motion tracking - as the method is based off of the method we used for diffeomorphic image registration, the algorithm will look very similar but will instead make use of the new closed form solutions. 

\begin{algorithm}[H]
	\caption{Motion Tracking} \label{alg:Motion Tracking}
	\textbf{Input Images Matrices:} $\vec{M}_0$ and $\vec{M}_1$\\
	\textbf{Input Parameters:} (levels, $N_{warp}$, $N_{iter}$, $\lambda$, $\theta$, $\alpha$, $\mu$, $\rho$, tolerance, difference)\\
	\For{\normalfont{\textbf{all}} $s \in$ levels}{
		$I_0 \gets resize^-(I_0, s)$\\
		$I_1 \gets resize^-(I_1, s)$\\
		\eIf{\normalfont$s = $ levels[$0$]}{
			$\vec{v}^0 = \vec{0}$
		}{
			$\vec{v}^s \gets resize^+(\vec{v}^s, 2)$
		}
		\While{\normalfont$\omega < N_{warp}$ or while Equation \ref{eq:Stopping Criteria Warp} is not true}{
			\If{$\omega = 0$}{
				$\vec{v}^{\omega} \gets \vec{v}^s$
			}
			$\vec{M}_1^{\omega} \gets warp(\vec{M}_1, \vec{v}^{\omega})$\\
			\textbf{Initialise:} $\vec{L}^0 = \vec{0}, \vec{S}^0 = \vec{0}, \vec{Y}^0 = \vec{0}, \vec{w}^0 = \vec{0}, \vec{b}^0 = \vec{0}$\\
			\While{\normalfont $k < N_{iter}$ or while Equation \ref{eq:Stopping Criteria Iter} is not true}{
				Update $\vec{L}^{k+1}$ with Equation \ref{eq:Motion Tracking L cases}\\
				Update $\vec{S}^{k+1}$ with Equation \ref{eq:Motion Tracking S cases}\\
				Update $\vec{v}^{k+1}$ with Equation \ref{eq:Motion Tracking v cases} and with $\vec{M}_1^{\omega}$\\
				Update $\hat{\vec{v}}^{k+1}$ with Equation \ref{eq:Motion Tracking v hat cases}\\
				Update $\vec{w}^{k+1}$ with Equation \ref{eq:Motion Tracking w cases}\\
				Update $\vec{Y}^{k+1}$ with equation \ref{eq:Motion Tracking Y cases}\\
				Update $\vec{b}^k$ with equation \ref{eq:Motion Tracking b cases}
			}
			$\vec{v}^{\omega} \gets \vec{v}^k$
		}
		$\vec{v}^s$ $\gets$ $\vec{v}^{\omega}$
	}
	\textbf{Return:} $\vec{v}^* \gets \vec{v}^s$
	
\end{algorithm} 

Algorithm \ref{alg:Motion Tracking} above, is the algorithm that we make use of for motion tracking, and is clearly very similar to Algorithm \ref{alg:Diffeomorphic Image Registration}. For this reason we will only describe parts of the algorithm that are different. For parts that are similar, we discussed them in Section \ref{sec:Diffeomorphic Image Registration}. \\

Here we have a few more parameters than before, namely $\mu$ and $\rho$. For this algorithm, we set $\mu = 0.2$ and $\rho = 0.1$. How these parameters were chosen is explained in section \ref{sec:Ablation Study}.\\

Furthermore, instead of having a source or target image we now instead have source and target matrices $\vec{M}_1$ and $\vec{M}_0$ and these were defined at the start of Section \ref{sec:Motion Tracking}. This means that we also have to make use of $\vec{M}$ (Equation \ref{eq:Casorati Matrix}) too which is the difference of these. Furthermore, instead of warping the source image we instead warp the source matrix to get $\vec{M}_1^\omega$.\\

Finally, we make use of the same stopping criteria that was used in Algorithm \ref{alg:Diffeomorphic Image Registration} for diffeomorphic image registration. These stopping criteria are Equations \ref{eq:Stopping Criteria Warp} and \ref{eq:Stopping Criteria Iter} and were discussed at the end of Section \ref{sec:Diffeomorphic Image Registration}. We do not make use of new stopping criteria or the stopping criteria for RPCA as our objective is still to obtain a deformation, $\vec{v}$, not to obtain a low rank matrix, $\vec{L}$. The objective of our algorithm is to obtain a useful, $\vec{v}$, such that we create a low rank matrix, $\vec{L}$, which is warped to the target. If $\vec{L}$ takes over the algorithm then we might obtain a low rank matrix which has not been warped properly. \\

Now that we have discussed the changes of this new algorithm to the diffeomorphic image registration algorithm, we can now look at the results and the outputs of the algorithm. 

\section{Results}
\label{sec:Motion Tracking Results}

The algorithm used for motion tracking is set out in Algorithm \ref{alg:Motion Tracking} and is used to do group-wise registration, however, it can be done on pairwise examples. The main advantages of our new algorithm for motion tracking is that it is designed to perform group-wise deformation on multiple images at once, something that Algorithm \ref{alg:Diffeomorphic Image Registration} is not designed for. Furthermore, the deformed groups of images will be low rank such that they all look the same; as this will mean they are aligned properly. For that reason, in this section we will discuss the advantages of the new algorithm only. In section \ref{sec:Algorithm Comparison}, we will do a comparison on pairwise deformations for this new approach to the old approach as the new approach should be able to do both to be successful. If we consider the new approach first (group-wise), then we must first have a sequence of images, and create our source and target matrices. Then we input these into Algorithm \ref{alg:Motion Tracking} and get our deformation as an output. Furthermore, we can also get a low rank matrix as an output too. However, in Section \ref{sec:Motion Tracking} we saw that we take a low rank and sparse decomposition of $\vec{M}$ (Equation \ref{eq:Motion Tracking Minimisation}), and this means that our low rank matrix will be a low rank representation of the difference between the source and target images - this is not what we want. In order to recover a low rank representation of the source images all we need to do is add the target matrix to this low rank matrix. Since the target matrix has columns which are linearly dependent, this will not change the rank of the low rank matrix at all - this allows us to visualise the low rank warped source images.\\

Since in Section \ref{sec:Diffeomorphic Image Registration Results} we went through simple examples of how the warping works, we will instead skip that in this section and just show complex results on the cardiac dataset. We will still explain the outputs and show multiple examples. 

\begin{figure}[H]
	\begin{center}
		\includegraphics[width = 0.9\textwidth]{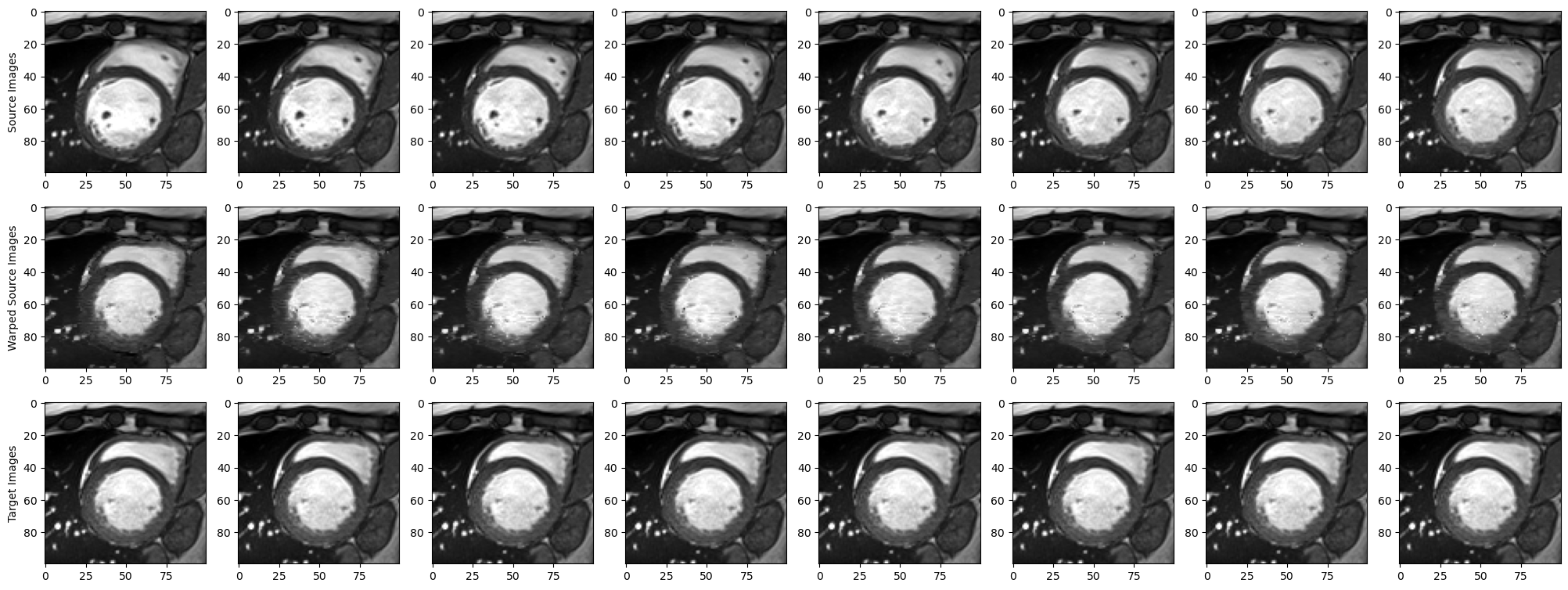}
		\caption{Motion Tracking Warping on 8 Frames Simultaneously} 
		\label{fig:Motion Tracking Patient001}
	\end{center}
\end{figure}

In Figure \ref{fig:Motion Tracking Patient001}, we see the effect of the output of the motion tracking Algorithm \ref{alg:Motion Tracking} on a sequence of cardiac images. Along the top row of Figure \ref{fig:Motion Tracking Patient001} we see the source images (for the algorithm these of course would be vectorised and concatenated into a matrix) which consist of a sequence of 8 frames of a cardiac cycle. Along the bottom row we see the target images which are all the same image and correspond to frame 11 of the cardiac cycle at ES. In the middle row of Figure \ref{fig:Motion Tracking Patient001} we see outputted deformation applied to the input images (since the input images are vectorised and put into a matrix, first we apply the deformation and then we must reshape the matrix back into individual images). We can look at the similarity of the middle row to the bottom row to try and determine the success of the algorithm. The main differences you can see from the top row (the sources) and the middle row (the warped sources) is that the size of the left ventricle (the circular part) is significantly decreased across all images. Furthermore, the shape of the left ventricle in the warped sources resembles that of the shape of the target image. Furthermore, the right ventricle has significantly decreased in size and looks similar to the right ventricle in the bottom row, this is particularly more noticeable on images towards the left hand side of Figure \ref{fig:Motion Tracking Patient001}. While this shows the output of the deformation when applied to the source images, we can also look at the low rank matrix as this is meant to be a low rank representation of the warping process. As previously discussed, due to our optimisation problem the low rank matrix will actually represent the difference between the warped source images and target images; in order to recover a low rank representation of the warped source images all we need to do it add the target matrix to the low rank matrix (we will call this the altered low rank matrix):

\begin{figure}[H]
	\begin{center}
		\includegraphics[width = 0.9\textwidth]{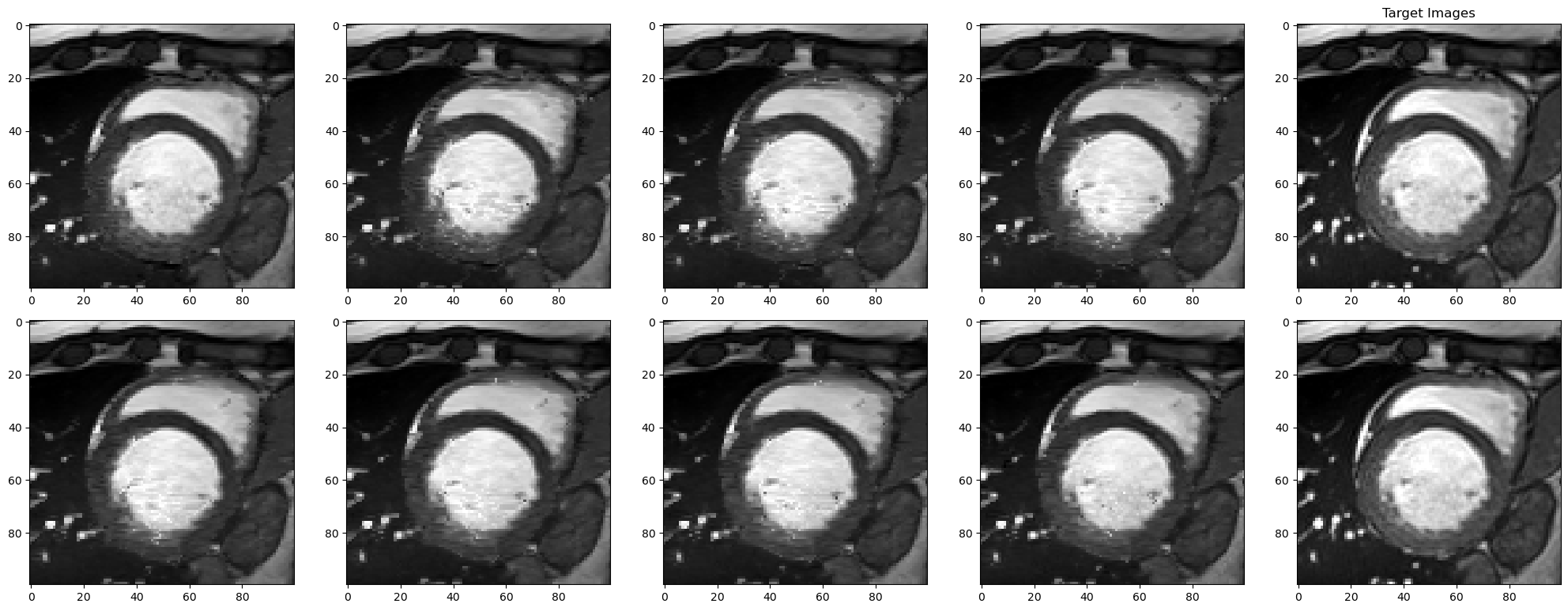}
		\caption{Motion Tracking Low Rank Warps Compared to Target} 
		\label{fig:Motion Tracking Patient001 Low Rank}
	\end{center}
\end{figure}

From Figure \ref{fig:Motion Tracking Patient001 Low Rank} we can see the altered low rank matrix in columns 1-4 which shows all 8 frames of the warped source images, while column 5 just shows the target image (both the same) for reference and comparison. As we see the warped source in columns 1-4 are low rank as they all look the same and are therefore linearly dependent within our low rank matrix $\vec{L}$ - this is an indicator that our registration is accurate.\\

Unfortunately, empirically testing the accuracy of these warpings becomes a challenge. As we did in Section \ref{sec:Diffeomorphic Image Registration Results}, we could do a warping between the source and target images and then apply the deformation to the ground truth source images. However, the dataset \cite{dataset} only has ground truth labels at ED and ES. The source images we used are from 8 frames within the cardiac cycle, 7 of these therefore do not have ground truth labels and so we cannot warp the ground truth labels and compare them to the target labels to calculate a Dice score. For this reason, we need to use another dataset which has the ground truth labels at every frame of the cardiac cycle, so we use the UK BioBank dataset \cite{biobank}. 

\begin{figure}[H]
	\centering
	\subfloat[Motion Tracking on 8 Frames Simultaneously (BioBank)]{\includegraphics[width=0.49\textwidth]{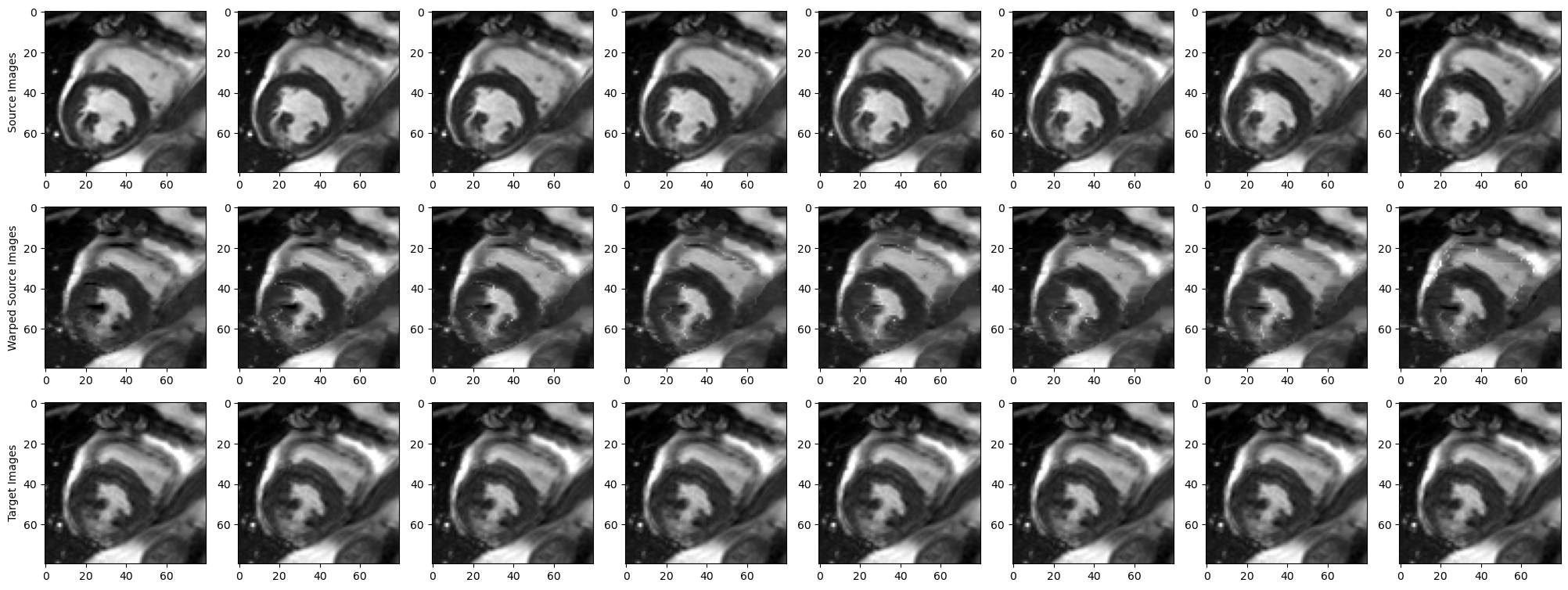}\label{fig:Motion Tracking BioBank Subplot}}
	\subfloat[Motion Tracking Warp Applied to GT labels]{\includegraphics[width=0.49\textwidth]{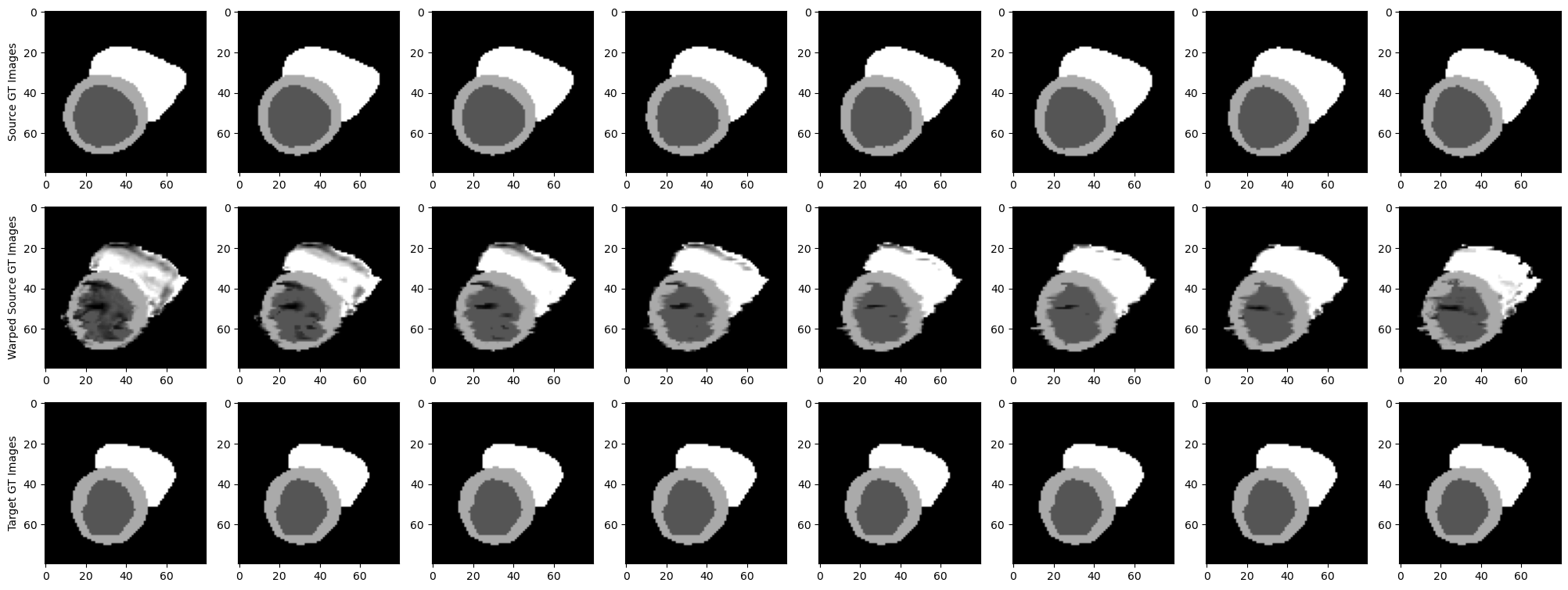}\label{fig:Motion Tracking BioBank GT Subplot}}
	
	\caption{Motion Tracking on a BioBank Dataset Example}
	\label{fig:Motion Tracking BioBank Complete Subplot}
\end{figure}

In Figure \ref{fig:Motion Tracking BioBank Complete Subplot}, we see two sub-figures, \ref{fig:Motion Tracking BioBank Subplot} shows the original images similar to Figure \ref{fig:Motion Tracking Patient001} where the top row is a sequence of 8 source images, the bottom row is the target image repeated, and the middle row shows the source image above it warped to look like the bottom target image. In Figure \ref{fig:Motion Tracking BioBank GT Subplot} we see a very similar sub-plot except on the ground truth labelled images, the top row of that figure showing the source ground truth images, the middle row showing those same images when applied with the deformation outputted from Algorithm \ref{alg:Motion Tracking}, and the bottom row showing the target ground truth images. As we can see, the middle row looks similar to the bottom from for Figure \ref{fig:Motion Tracking BioBank GT Subplot} - this is again noticeable by the decreased size of the left ventricle when compared to the source images (the dark grey), and the decreased size of the right ventricle too (the white). We do unfortunately now notice the slight inaccuracies in terms of noise, particularly in the images of the left of the middle row. This is a side effect of vectorising the images into matrices, where the deformation sometimes has a tendency to warp into neighbouring images creating this slight inaccuracy. Fortunately, now that we have ground truth labels we can measure the Dice coefficient for every single frame and take an average to get a measure of success for the group-wise deformation. Using Equation \ref{eq:Dice} we get a dice score for the 8 frames in Figure \ref{fig:Motion Tracking BioBank Complete Subplot} of 0.7765. This score is pretty good so afterwards we tested the same heart but instead of using the first 8 frames of the cardiac cycle, we instead used the first 16 - this test returned a dice score of 0.8535. As previously stated, testing on a single example cannot be used to say with certainty whether an algorithm is successful - however, due to limitations with the amount of data, we are unable to test this example any further. For this reason we can only subjectively argue whether this algorithm is a success - to aide with this we will observe this algorithm applied to 3 more examples. Each of these is done the same as the example above and the warping is done on 16 frames simultaneously, however, only 8 of them will be visualised. 

\begin{figure}[H]
	\centering
	\subfloat{\includegraphics[width=0.49\textwidth]{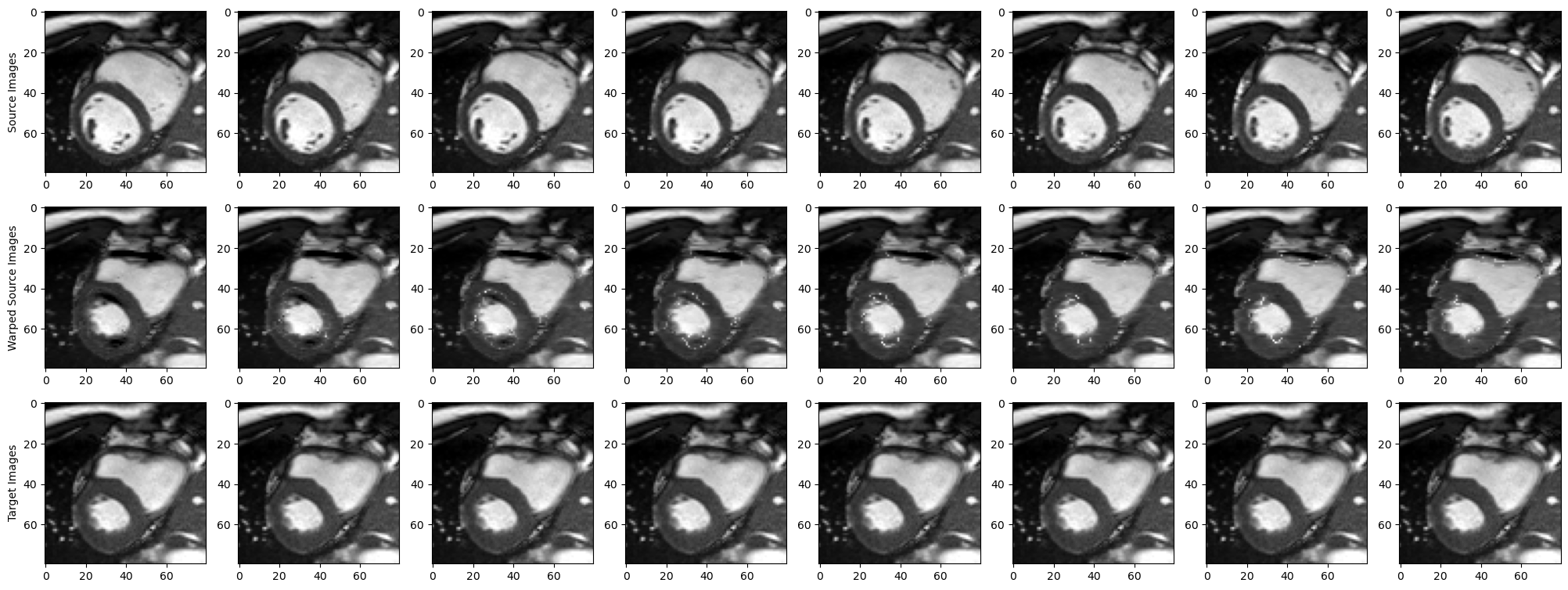}\label{fig:Motion Tracking BioBank 2}}
	\subfloat{\includegraphics[width=0.49\textwidth]{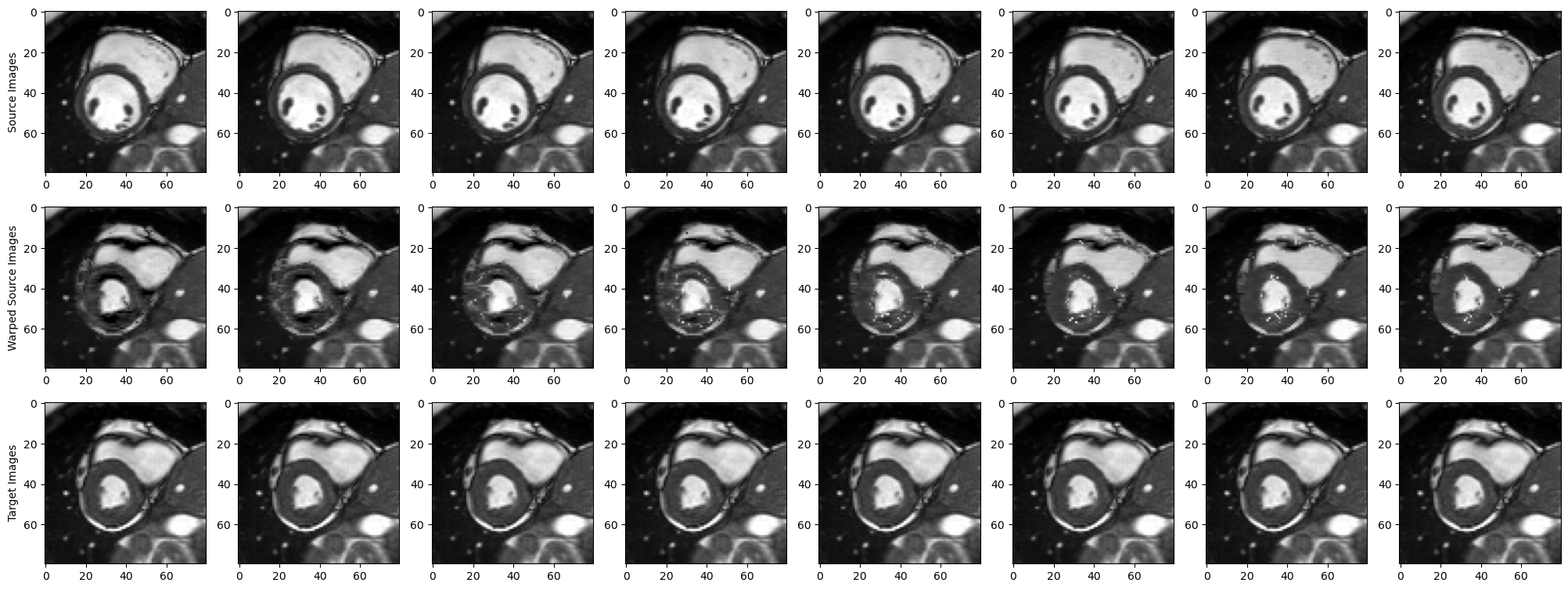}\label{fig:Motion Tracking BioBank GT 3}}
	\hfill
	\subfloat{\includegraphics[width=0.49\textwidth]{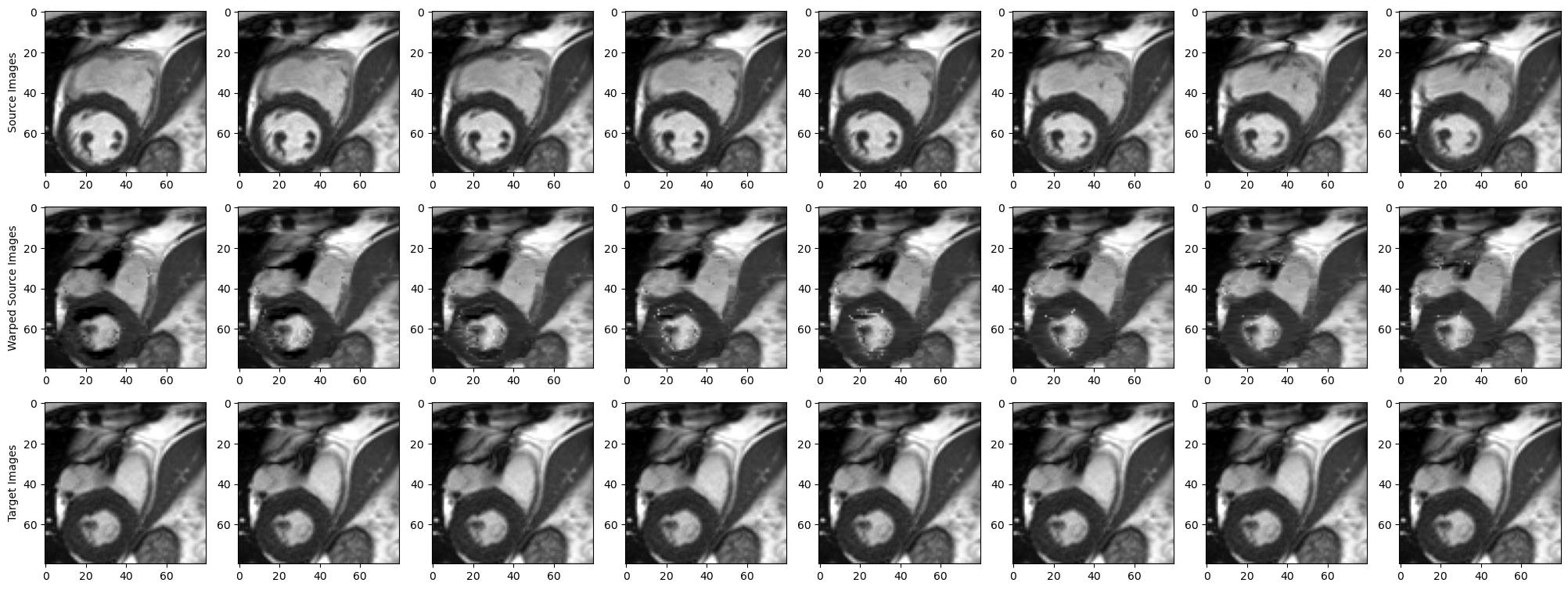}\label{fig:Motion Tracking BioBank GT 4}}
	
	\caption{Motion Tracking on The Remaining BioBank Dataset Examples}
	\label{fig:Motion Tracking BioBank Remaining}
\end{figure}

While we can look at these subjectively and comment about how the algorithm looks to be successful, it is hard to say without a deeper analysis (an analysis and comparison on a further 16 examples is provided in Section \ref{sec:Algorithm Comparison}). The Dice coefficient on these other 3 examples were 0.8576, 0.7459, 0.7992, which are very good scores and show that the algorithm can be successful. 

\section{Ablation Study}
\label{sec:Ablation Study}

Furthermore, we did an in-depth analysis on the new motion tracking parameters $\mu$ and $\rho$ to try and determine the optimal value to be used. The parameters which are carried across from Algorithm \ref{alg:Diffeomorphic Image Registration} ($\theta$, $\lambda$, $\alpha$) are kept the same because in \cite{thorley2021nesterov} they did their own deep analysis to obtain the optimal parameters.

\begin{figure}[H]
	\begin{center}
		\includegraphics[width = 0.9\textwidth]{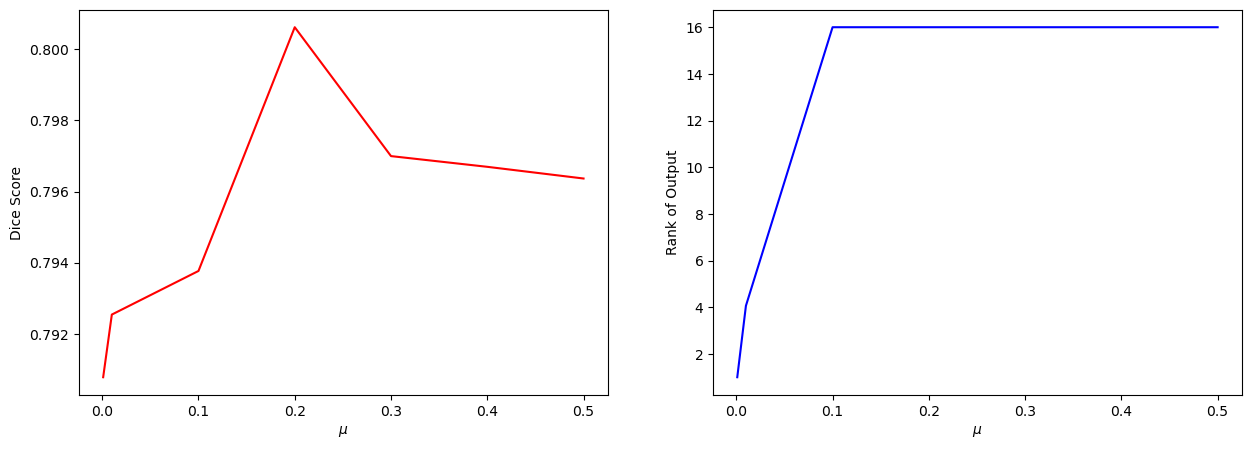}
		\caption{Optimal $\mu$ for The Motion Tracking Algorithm \ref{alg:Motion Tracking}} 
		\label{fig:Motion Tracking Optimal mu}
	\end{center}
\end{figure}

In the above Figure \ref{fig:Motion Tracking Optimal mu}, we see two figures, the one on the left shows how varying $\mu$ leads to a change in Dice score. The figure on the right shows how varying $\mu$ leads to a change in the rank of the outputted warped images. For this experiment, we tested the motion tracking algorithm on 20 examples and took the mean of the dice score or rank across all 20 examples for a given $\mu$. The values of $\mu$ that were tested for both experiments were: $0.001, 0.01, 0.1, 0.2, 0.3, 0.4, 0.5$. We can draw multiple conclusions from Figure \ref{fig:Motion Tracking Optimal mu}. Firstly, if we are only concerned with an optimal dice score then $\mu = 0.2$ is the best value. This values gives us the largest dice score and therefore the most accurate group-wise registration. Secondly, if we are concerned with a low rank warp, then a $\mu \geq 0.1$ will never produce this. To decrease the rank of the warp, we need to have a $\mu < 0.1$. Thirdly, there is a trade-off between the rank of the output and accuracy. The lower the $\mu$ then the lower the rank of the output, however, the algorithm will focus so much on decreasing the rank that it will sacrifice some precision. We see that when $\mu = 0.01$ we have a rank of around $4$ but a dice score greater than when $\mu = 0.001$. However, when $\mu = 0.001$ we will create warped images with a rank of $1$. For this reason, $\mu$ should never be less than $0.001$. Due to the information in the above Figure \ref{fig:Motion Tracking Optimal mu} we set $\mu = 0.2$ if we want to focus on precision, and $\mu = 0.01$ if we desire low rank warps. Below is a summary of this information, with a figure to help.\\

If we set $\mu = 0.01$ then we get a similar dice score on several examples and we do achieve a low rank matrix, not just based off of looks, but if we test the rank of the low rank output, $\vec{L}$, then it will have a rank around 5 even though we have 16 input images. 

\begin{figure}[H]
	\begin{center}
		\includegraphics[width = 0.9\textwidth]{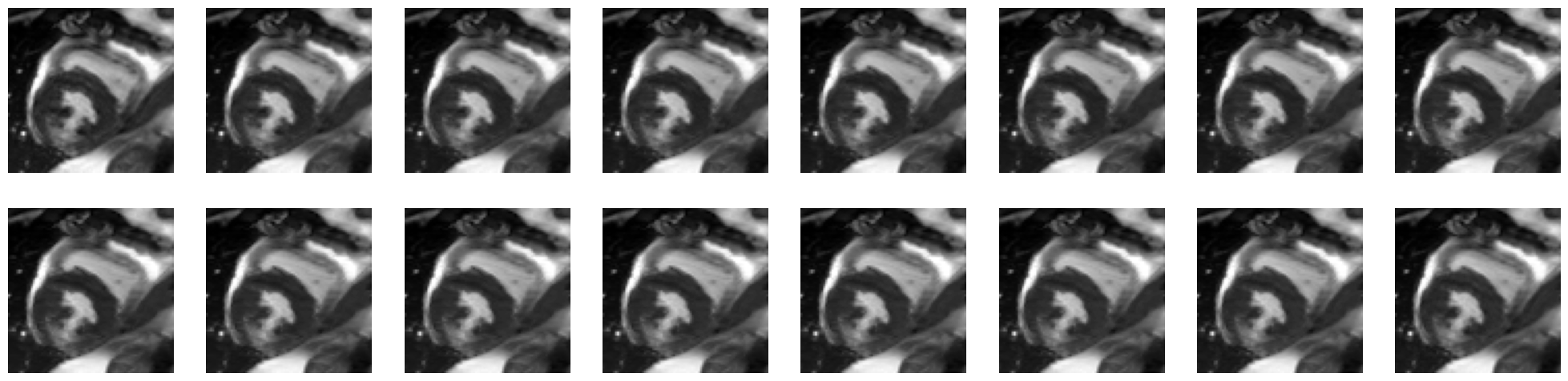}
		\caption{Motion Tracking Altered Low Rank Output} 
		\label{fig:Motion Tracking Rank 5 Image}
	\end{center}
\end{figure}

The above Figure \ref{fig:Motion Tracking Rank 5 Image}, is one such example of this where the low rank output has a rank of 5, and we can see how similar all the warped images look. This proves that the images are being warped such that they look the same as it decreases the rank from 16 linearly independent images to only 5. These 5 independent images might not be warped to look like the target image though. To check this we can concatenate the target image to the low rank matrix, $\vec{L}$, and see if the rank increases (because the target image doesn't appear in $\vec{L}$) or if it stays the same (because the warped images look exactly like the target image). When we do this we notice that the rank stays the same. This is excellent as it shows that the motion tracking Algorithm \ref{alg:Motion Tracking} provides accurate low rank warps when performed on group-wise registration. Finally, if we set $\mu = 0.005$ then this parameter will really try and force our algorithm to make a low rank warp. When we do this, we get competitive dice scores of around $0.83$, however, the rank of our output is $1$ clearly demonstrating the ability of the algorithm to create a low rank output. \\

\begin{figure}[H]
	\begin{center}
		\includegraphics[width = 0.9\textwidth]{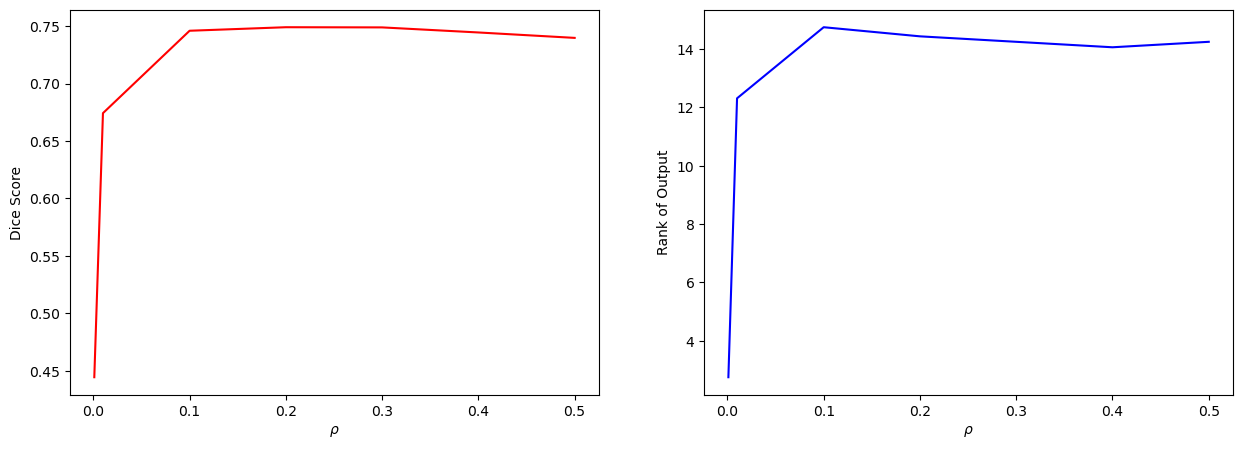}
		\caption{Optimal $\rho$ for The Motion Tracking Algorithm \ref{alg:Motion Tracking}} 
		\label{fig:Motion Tracking Optimal rho}
	\end{center}
\end{figure}

In Figure \ref{fig:Motion Tracking Optimal rho} above, we again see two figures, the one on the left shows how a varying $\rho$ affects the dice score of the output of the motion tracking algorithm \ref{alg:Motion Tracking}, while the figure on the right shows how the varying of $\rho$ affects the rank of the low rank matrix $\vec{L}$. For this study, we again tested on 20 different group-wise registration examples, where we attempt to warp 16 images together simultaneously. The values of $\rho$ that were tested were: $0.001, 0.01, 0.1, 0.2, 0.3, 0.4$ and $0.5$. We can draw several conclusions from Figure \ref{fig:Motion Tracking Optimal rho}, firstly, the optimal value for $\rho$ is $0.1$. This is evident as we see that it produces the highest dice score, however, it is also evident that higher values of $\rho$ produce competitive dice scores too. We also see that a $\rho$ of $0.01$ reduces the outputted dice score by about $0.05$, but, from the figure on the right of Figure \ref{fig:Motion Tracking Optimal rho} we see that it does reduce the rank of the output. Finally, we see that a really low value of $\rho$ like $0.001$ produces very low rank outputs, however, they have terrible dice scores and therefore are not accurate warps. \\

Using Figures \ref{fig:Motion Tracking Optimal mu} and \ref{fig:Motion Tracking Optimal rho}, we can now determine the optimal values for our hyperparameters. For the most accurate group-wise registration we need to choose: $\mu = 0.2, \rho=0.1$. These values produce group-wise registrations with the highest dice scores possible and therefore produce the most accurate registrations. If we would like to produce a low rank registration, then we should not touch the value for $\rho$ as we clearly see from Figure \ref{fig:Motion Tracking Optimal rho} that $\rho$ is the most sensitive out of the two parameters, and decreasing the value for the parameter, decreases the outputted dice score too much. For this reason, for low rank group-wise registrations we should choose: $\mu = \{0.01, 0.001\}, \rho=0.1$. This way we will produce a low rank group-wise registration while maintaining the highest dice score possible. Producing low rank group-wise registrations was our aim, therefore, these parameters achieve this. \\ 

Finally, it is worth noting that in Figures \ref{fig:Motion Tracking Optimal mu} and \ref{fig:Motion Tracking Optimal rho}, the values for the dice scores are lower than what we state in our results, because to find the optimal parameters we lowered the stopping criteria (Equations \ref{eq:Stopping Criteria Iter}, \ref{eq:Stopping Criteria Warp}) so that the algorithm would run faster. \\

As stated previously, we wanted to acknowledge the new ideas that this algorithm brought in this section. As we have seen in this section our motion tracking algorithm can perform group-wise deformation with good accuracy and also create low rank outputs such that all warped images look the same. This is exactly what we want to track the motion of the heart, we need to know how much each section of the heart is moving by in each frame, and if we can align all frames of the heart to look identical, then we know exactly how much each part of the heart has moved by. \\

With this revelation, we will now move on to compare this new algorithm to the old diffeomorphic image registration algorithm and come to conclusions about our success within our task. 

\chapter{Conclusion}
\label{cha:Conclusion}

\section{Algorithm Comparison}
\label{sec:Algorithm Comparison}

In Section \ref{sec:Motion Tracking Results} we discussed the advanatages and what it was that was new that the motion tracking algorithm could achieve. We mentioned how it was accurate and provided good results for group-wise deformation which is essential to track the motion of the heart - also it is designed for group-wise deformations whereas diffeomorphic image registration was not. Furthermore, we saw that we could create a low rank warped output such that each frame in the output looked identical. While this is excellent, it should still be able to perform well when compared against diffeomorphic image registration. For the first example we will simply compare on patient001 as we had a look at this deformation in Section \ref{sec:Diffeomorphic Image Registration Results}. 

\begin{figure}[H]
	\centering
	\subfloat[DiffIR]{\includegraphics[width=0.49\textwidth]{Images/diffir_patient001_example}\label{fig:DiffIR Patient001 Comparison}}
	\subfloat[Motion Tracking]{\includegraphics[width=0.49\textwidth]{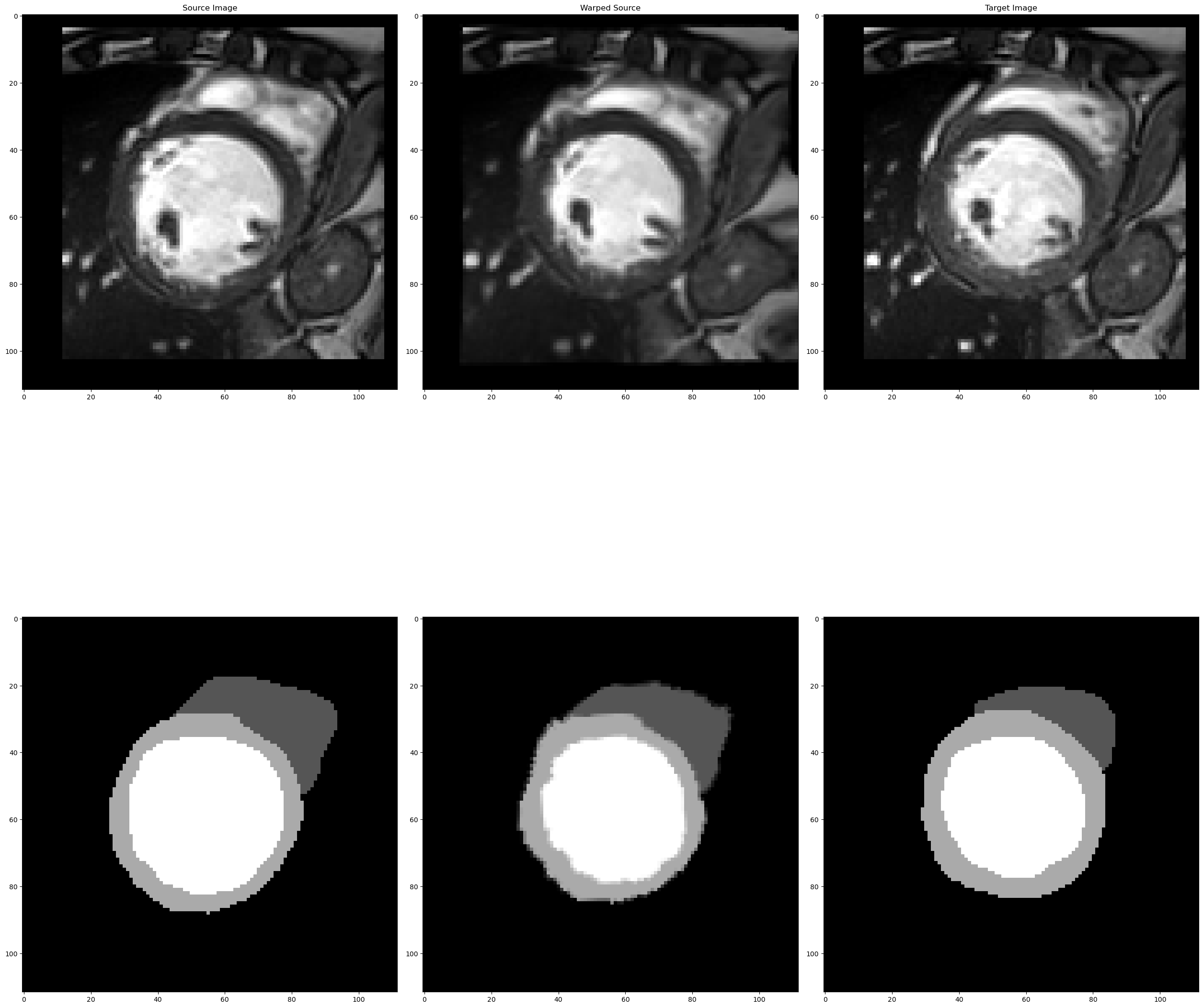}\label{fig:Motion Patient001 Comparison}}
	\hfill
	\subfloat[DiffIR]{\includegraphics[width=0.49\textwidth]{Images/diffir_patient001_example_deformation}\label{fig:DiffIR Patient001 Comparison Deformation}}
	\subfloat[Motion Tracking]{\includegraphics[width=0.49\textwidth]{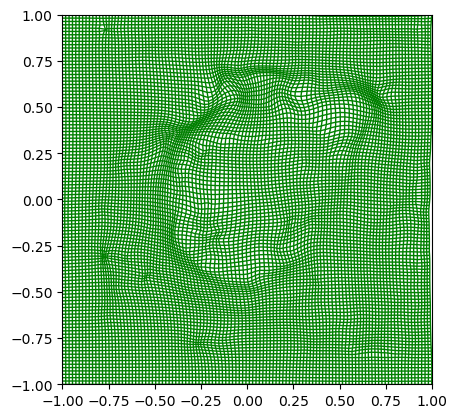}\label{fig:Motion Patient001 Comparison Deformation}}
	\hfill
	\subfloat[DiffIR]{\includegraphics[width=0.49\textwidth]{Images/diffir_patient001_example_hsv}\label{fig:DiffIR Patient001 Comparison HSV}}
	\subfloat[Motion Tracking]{\includegraphics[width=0.49\textwidth]{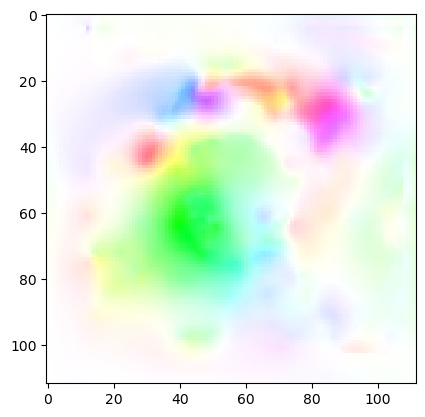}\label{fig:Motion Patient001 Comparison HSV}}
	
	\caption{DiffIR vs Motion Patient001}
	\label{fig:Patient001 Comparison}
\end{figure}

Here we see that the warping is very similar for both diffeomorphic image registration and motion tracking. This is paralleled when we compare the dice coefficient for both examples, where motion tracking scored 0.8650 and DiffIR scored 0.8621, so a very minor increase. We can also compare the deformation fields where the motion tracking field looks a bit smoother particularly around the right ventricle area, however, it also unnecessarily warps the edges of the image. This is easily portrayed in the HSV images where the motion tracking HSV, Figure \ref{fig:Motion Patient001 Comparison HSV}, has softer colours and then has unnecessary colours which are quite dark around the edges. 

\begin{figure}[H]
	\centering
	\subfloat{\includegraphics[width=0.98\textwidth]{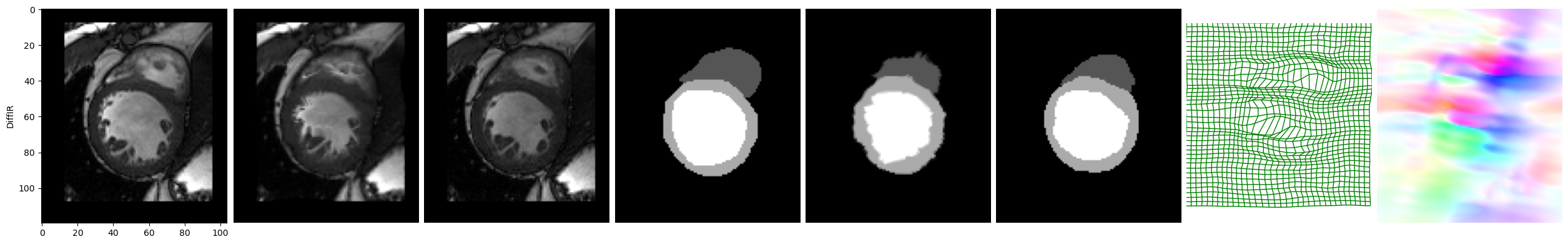}}
	\vspace*{-5mm}
	\subfloat{\includegraphics[width=0.98\textwidth]{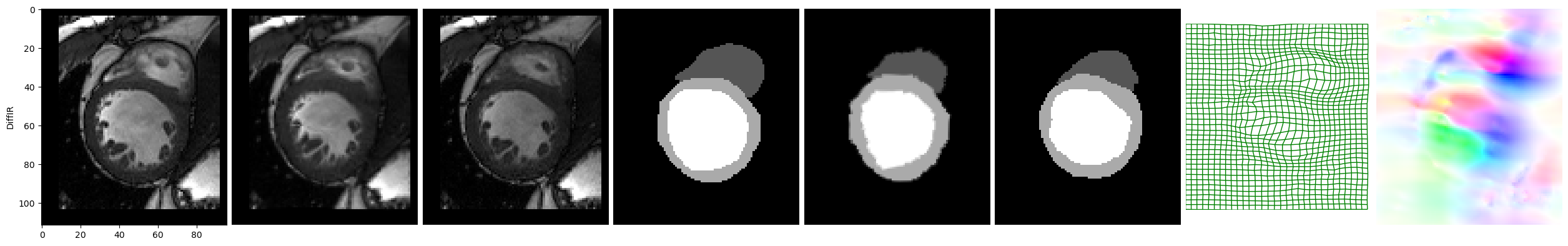}}
	\vspace*{-5mm}
	\subfloat{\includegraphics[width=0.98\textwidth]{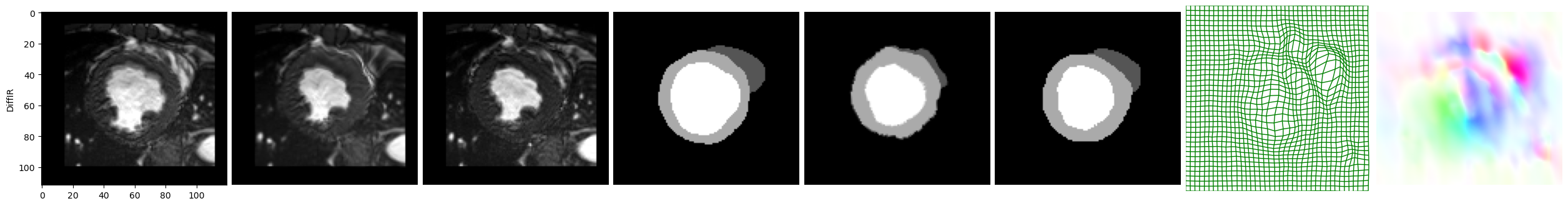}}
	\vspace*{-5mm}
	\subfloat{\includegraphics[width=0.98\textwidth]{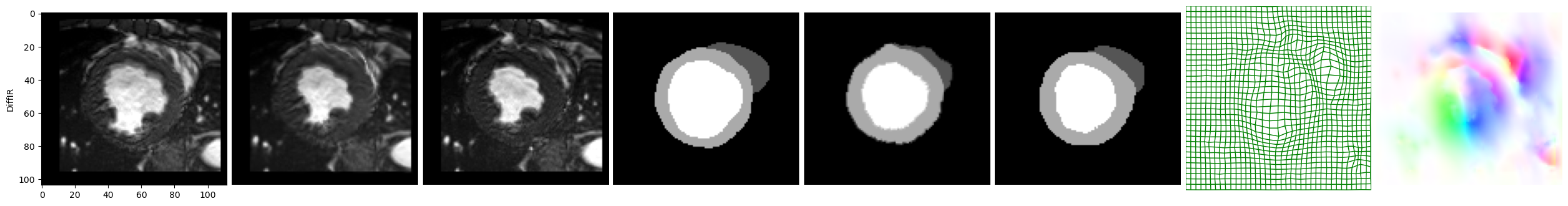}}
	\vspace*{-5mm}
	\subfloat{\includegraphics[width=0.98\textwidth]{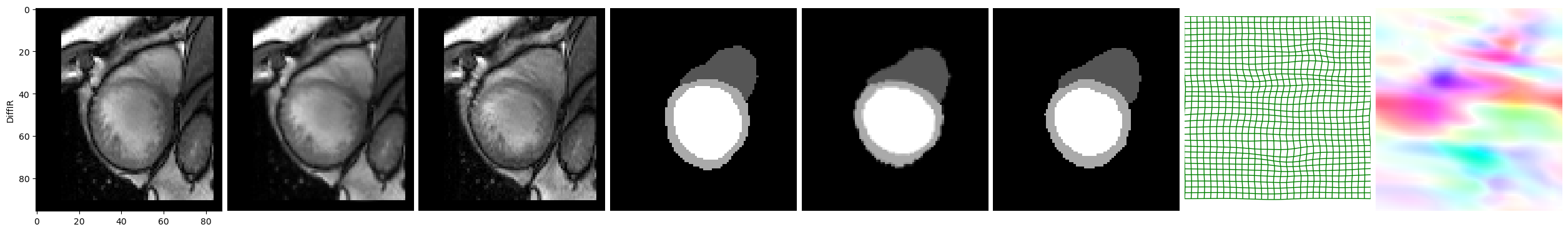}}
	\vspace*{-5mm}
	\subfloat{\includegraphics[width=0.98\textwidth]{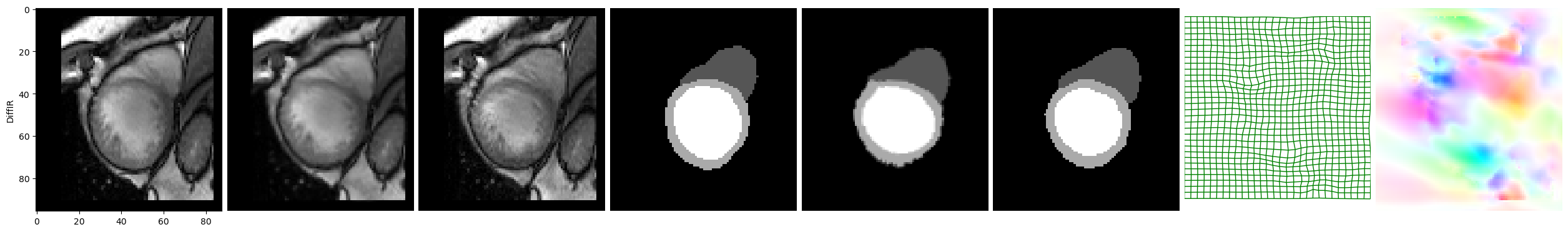}}
	\caption{Multiple Comparisons of DiffIR vs Motion Tracking}
	\label{fig:DiffIR Vs Motion}
\end{figure}

In the above Figure \ref{fig:DiffIR Vs Motion} we see a few comparisons between DiffIR and motion tracking in pairwise registration. Every 2 rows we have one example from DiffIR and the other is from motion tracking, on the same patient. For each column going from left to right we have; the source image; the warped source; the target image; the ground truth source; the deformation applied to the ground truth source; the target image's ground truth; the deformation field and finally the HSV representation of the deformation field. In the first example in rows 1 and 2, motion tracking performs slightly better - and this is seen by the better shape of the right ventricle, and the left ventricle being more appropriately warped. The motion tracking algorithm had a dice coefficient of 0.8400 while DiffIR had one of 0.8087, for that patient. In the second example, in rows 3 and 4 we see motion tracking warping slightly better again - while the left ventricle is warped well in both cases, the motion tracking algorithm seems to have not over warped the right ventricle and for this reason its dice coefficient is 0.8391 while DiffIR has one of 0.8020. In the final example in rows 5 and 6 we see DiffIR outperforming the motion tracking algorithm as the motion tracking has a slightly less accurate warping around the edges of the right ventricle and in some areas of the left ventricle too. For this reason DiffIR has a dice score of 0.8629 while the motion tracking algorithm had one of 0.8569.\\

These examples are slightly subjective and they will help guide us to see that both algorithms are good choice when it comes to doing pairwise registration. In Section \ref{sec:Diffeomorphic Image Registration Results}, we discussed how we tested the diffeomorphic image registration Algorithm \ref{alg:Diffeomorphic Image Registration} on 100 examples to get an average dice score of 0.8170. We can do the exact same thing for the motion tracking Algorithm \ref{alg:Motion Tracking}. A test on 100 examples shows an average dice score of $0.8107\pm0.068$.

\begin{figure}[H]
	\centering
	\subfloat{\includegraphics[width=0.49\textwidth]{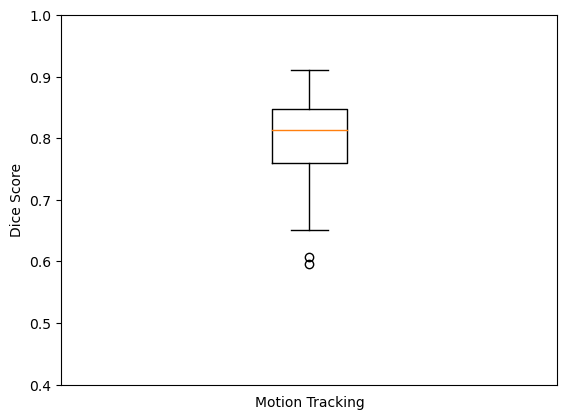}}
	\subfloat{\includegraphics[width=0.49\textwidth]{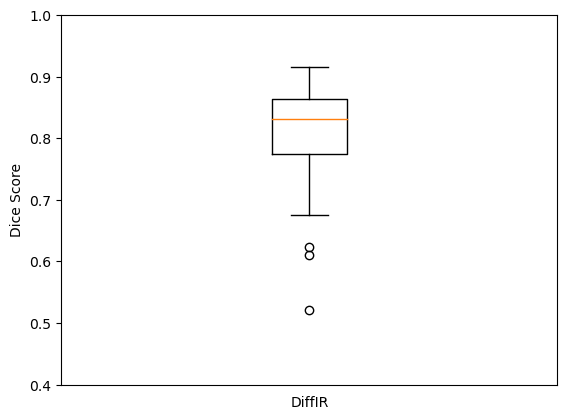}}
	\caption{Box plots of Dice Scores for Motion Tracking and DiffIR on Pairwise Registration}
	\label{fig:Pairwise Boxplots}
\end{figure}

This would imply that the motion tracking, for pairwise registration, is slightly worse, by a score of 0.0063 (DiffIR is better by $0.78\%$). Figure \ref{fig:Pairwise Boxplots} shows a box plot comparison for the motion tracking Algorithm \ref{alg:Motion Tracking} and the DiffIR Algorithm \ref{alg:Diffeomorphic Image Registration} on 100 pairwise examples. This shows how competitive the motion tracking algorithm is to the DiffIR algorithm as they have very similar metrics on the box plots. The two algorithms have very similar means, interquartile ranges, and ranges. However, from Figure \ref{fig:DiffIR Vs Motion} we see that the motion tracking algorithm can in some situations be either better or worse, so this would imply that the motion tracking algorithm is competitive with the diffeomorphic image registration algorithm and that is very good. It is vital to remember that the motion tracking algorithm is not centred around pairwise registration, but group-wise registration, and that is where the strong advantages of the algorithm lie. The main upsides of the motion tracking algorithm are with its ability to provide accurate registrations between multiple images at once. \\

To further this, we will try the diffeomorphic image registration on the same group-wise examples that we did in Section \ref{sec:Motion Tracking Results}. For those 4 examples, motion tracking provided dice scores of: 0.8535, 0.8576, 0.7459, 0.7992. On the same examples, diffeomorphic image registration provides a score of: 0.8266, 0.8038, 0.5169, 0.7455. These results show that the motion tracking algorithm provides a more accurate registration on all examples when doing a group-wise registration. A more thorough analysis can be carried out if we use the same 4 examples that we used in Section \ref{sec:Motion Tracking Results}. If we use those same 4 cardiac sequences with 16 frames, but instead we use a different slice along the short axis then we can create 16 more examples along with the original 4 that can be used for analysis. Across all 20 examples diffeomorphic image registration when doing group-wise deformation on 16 frames at once had a mean dice score of 0.7352, whereas across the same 20 examples with the same stopping criteria and parameters, the motion tracking algorithm had a mean dice score of 0.8422, an increase of 0.107 or approximately $14.6\%$.

\begin{figure}[H]
	\centering
	\subfloat{\includegraphics[width=0.49\textwidth]{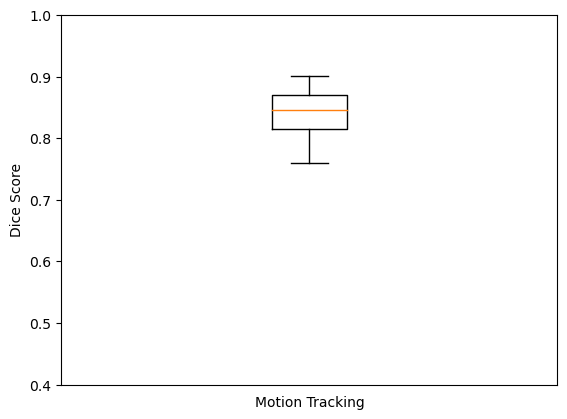}}
	\subfloat{\includegraphics[width=0.49\textwidth]{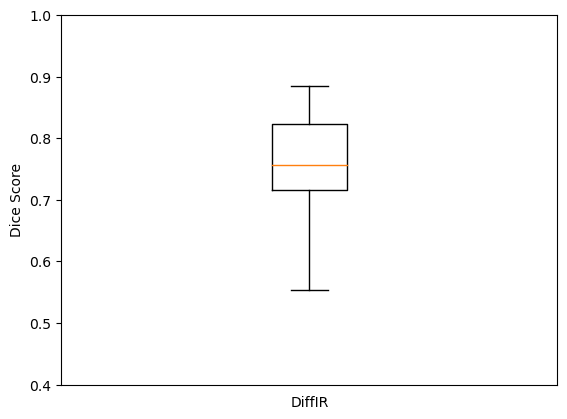}}
	\caption{Box plots of Dice Scores for Motion Tracking and DiffIR on Group-wise Registration}
	\label{fig:Groupwise Boxplots}
\end{figure}

From Figure \ref{fig:Groupwise Boxplots} we can see how the motion tracking Algorithm \ref{alg:Motion Tracking} outperforms the diffeomorphic image registration Algorithm \ref{alg:Diffeomorphic Image Registration}. This is evident and diffeomorphic image registration has a much larger range and a slightly larger interquartile range. Furthermore, the motion tracking algorithm has a much higher mean, and the dice scores for it's interquartile range are higher too. Finally, we can compare run times for each algorithm. Both of these were run on a CPU not a GPU. The motion tracking algorithm had a runtime around 60 seconds whereas diffeomorphic image registration algorithm was around 15 seconds. While one could suggest that the diffeomorphic image registration algorithm would be more competitive to the motion tracking algorithm if it was run for a similar time (they were run with the same convergence parameters) in experiments we found that decreasing the convergence parameters for diffeomorphic image registration so it could also run for 60 seconds actually made it less accurate.

\begin{table}[H]
	\centering
	\begin{tabular}{ | m{5em} | m{3cm}| m{3cm} | }
		\hline
		 & Motion Tracking & DiffIR \\
		\hline
		Group-wise (20) & $0.8422\pm0.040$ & $0.7352\pm0.094$\\
		\hline
		Pairwise (100) & $0.8107\pm0.068$ & $0.8170\pm0.072$\\
		\hline
		Group-wise Runtime (10) & 67.19 seconds & 11.05 seconds \\
		\hline
		Pairwise Runtime (10) & 26.99 seconds & 1.74 seconds \\
		\hline
	\end{tabular}
	\caption{Table of Comparisons between Motion Tracking algorithm \ref{alg:Motion Tracking} and DiffIR algorithm \ref{alg:Diffeomorphic Image Registration}}
	\label{table:Motion Tracking vs DiffIR}
\end{table}

In Table \ref{table:Motion Tracking vs DiffIR} above, we see a comparison between the motion tracking Algorithm \ref{alg:Motion Tracking} and the diffeomorphic image registration Algorithm \ref{alg:Diffeomorphic Image Registration}. The left hand column shows different objectives that the algorithms will be compared with, and the number in brackets shows the number of examples that were used to calculate the scores in the table. In the top 2 rows of the table, we see the dice scores on group-wise and pairwise image registration with standard deviations shown, and in the bottom 2 we see a comparison in runtimes for each example when run on a CPU not a GPU. This briefly summarises the information we discussed earlier in this chapter and affirms our conclusion that our motion tracking algorithm is better for equipped for low rank group-wise image registration. 

\section{Conclusion}
\label{sec:Conclusion}

In Section \ref{sec:Motivations and Aims}, we stated how our main aim is to create an algorithm that can perform motion tracking such that it can do group-wise deformations. Furthermore, we desire for this motion tracking algorithm to output warped images that are low rank as this would demonstrate that our warpings are accurate. To this end, we presented our motion tracking Algorithm \ref{alg:Motion Tracking}, and stated in Section \ref{sec:Motion Tracking Results} how the algorithm itself can perform group-wise deformations that will be low rank. We can conclude that our algorithm is successful as we have achieved all that we aimed for. In group-wise deformation examples, we obtained a mean dice score of $0.8422$, furthermore, we created low rank deformations decreasing the rank of the warped source images from $16$ to $5$, and, with a sufficiently small enough $\mu$, to a rank of $1$. The current state-of-the-art diffeomorphic image registration algorithm has a dice score of $0.7352$ on these examples, and does not produce a low rank output. This demonstrates the success of our algorithm proving that it can track the motion of the heart, being able to warp multiple images to look exactly like one other image and increasing the dice score of the previous algorithm by 14.6$\%$. Furthermore, the motion tracking algorithm displayed an increased dice score even if both algorithms were run for a similar amount of time or to the same convergence parameters. This is exactly what we aimed to achieve, since clearly we can now accurately track points across a sequence of images since we know how to deform all of the images to look exactly like a target image.

\addcontentsline{toc}{chapter}{References}
\renewcommand\bibname{References}
\printbibliography

@article{horn_schunck,
	author = {Schunck, B.G. and Horn B.K.}, 
	title = {Determining optical flow.}, 
	journal = {Artificial Intelligence 17},
	year = {1981},
}

@article{fastdiffeoIRalgorithm,
	author = {J, Ashburner}, 
	title = {A fast diffeomorphic image registration algorithm}, 
	journal = {NeuroImage 38},
	year = {2007},
}

@article{thorley2021nesterov,
	title={Nesterov Accelerated ADMM for Fast Diffeomorphic Image Registration}, 
	author={Alexander Thorley and Xi Jia and Hyung Jin Chang and Boyang Liu and Karina Bunting and Victoria Stoll and Antonio de Marvao and Declan P. O'Regan and Georgios Gkoutos and Dipak Kotecha and Jinming Duan},
	year={2021},
	journal={Medical Image Computing and Computer Assisted Intervention},
	pages={150-160},
}

@article{diffeoIReulerexample,
	author = {J., Ashburner and K. J. Friston}, 
	title = {Diffeomorphic registration using geodesic shooting and Gauss–Newton optimisation}, 
	journal = {NeuroImage},
	year = {2011},
}

@article{arbitraryordertotalvariation,
	title = {Arbitrary Order Total Variation for Deformable Image Registration},
	journal = {Pattern Recognition},
	volume = {137},
	pages = {109318},
	year = {2023},
	issn = {0031-3203},
	doi = {https://doi.org/10.1016/j.patcog.2023.109318},
	url = {https://www.sciencedirect.com/science/article/pii/S0031320323000195},
	author = {Jinming Duan and Xi Jia and Joseph Bartlett and Wenqi Lu and Zhaowen Qiu},
}

@book{admm,
	author={Boyd, Stephen and Parikh, Neal and Chu, Eric and Peleato, Borja and Eckstein, Jonathan},
	title={Distributed Optimization and Statistical Learning via the Alternating Direction Method of Multipliers},
	year={2011}
}

@article{nesterovaccelerated,
	title = {A method of solving a convex programming problem with convergence rate O(1/k2)},
	journal = {Doklady Akademii Nauk},
	volume = {269},
	pages = {543-547},
	year = {1983},
	author = {Nesterov, Y.E.},
}

@article{diffeodemons,
	title = {Diffeomorphic demons: Efficient non-parametric image registration},
	journal = {NeuroImage},
	volume = {45},
	number = {1, Supplement 1},
	pages = {S61-S72},
	year = {2009},
	note = {Mathematics in Brain Imaging},
	issn = {1053-8119},
	doi = {https://doi.org/10.1016/j.neuroimage.2008.10.040},
	url = {https://www.sciencedirect.com/science/article/pii/S1053811908011683},
	author = {Tom Vercauteren and Xavier Pennec and Aymeric Perchant and Nicholas Ayache},
}

@article{chen2020anatomyaware,
	title={Anatomy-Aware Cardiac Motion Estimation}, 
	author={Pingjun Chen and Xiao Chen and Eric Z. Chen and Hanchao Yu and Terrence Chen and Shanhui Sun},
	journal={Machine Learning in Medical Imaging},
	year={2020},
}

@article{2019siameseconvolution,
	author={Tang, Jiangyue and Gan, Ziyu and Yang, Xuan},
	journal={IEEE International Conference on Bioinformatics and Biomedicine (BIBM)}, 
	title={Cardiac Motion Tracking in Short-axis MRI using Siamese Convolution Network}, 
	year={2019},
	volume={},
	number={},
	pages={865-870},
	doi={10.1109/BIBM47256.2019.8982995}}

@article{candes2009robust,
	title={Robust Principal Component Analysis?}, 
	author={Cand\`{e}s, Emmanuel J. and Li, Xiaodong and Ma, Yi and Wright, John},
	year={2009},
	journal={Association for Computing Machinery},
	volume = {58},
	number = {3},
}

@article{lucaskanade,
	TITLE = {{An Iterative Image Registration Technique with an Application to Stereo Vision}},
	AUTHOR = {Lucas, Bruce D and Kanade, Takeo},
	URL = {https://hal.science/hal-03697340},
	journal = {{IJCAI'81: 7th international joint conference on Artificial intelligence}},
	ADDRESS = {Vancouver, Canada},
	VOLUME = {2},
	PAGES = {674-679},
	YEAR = {1981},
	HAL_ID = {hal-03697340},
	HAL_VERSION = {v1},
}

@article{lucaskanadeanalysis,
	title={Optimal Filter Estimation for Lucas-Kanade Optical Flow.}, 
	author={Sharmin N, Brad R.},
	year={2012},
	journal={Sensors (Basel, Switzerland)},
	volume={12},
	pages={12694-709},
}

@article{LDDMM,
	title={Computing Large Deformation Metric Mappings via Geodesic Flows of Diffeomorphisms.},
	author={Beg, M.F. and Miller, M.I. and Trouve, A. and Younes, L.},
	year={2005},
	pages={139-157},
	journal={International Journal of
	Computer Vision}
}

@article{FLASH,
	title={ Fast diffeomorphic image registration via fourierapproximated lie algebras.},
	author={Zhang, M. and Fletcher, P.T},
	year={2019},
	pages={61-73},
	journal={International Journal of Computer Vision}
}

@article{haase2020deformable,
	title={Deformable Groupwise Image Registration using Low-Rank and Sparse Decomposition}, 
	author={Roland Haase and Stefan Heldmann and Jan Lellmann},
	year={2022},
	journal={Journal of Mathematical Imaging and Vision},
	volume={64},
	pages={194-211},
}

@article{cai2008singular,
	title = {A Singular Value Thresholding Algorithm for Matrix Completion},
	author = {Cai, Jian-Feng and Cand\`{e}s, Emmanuel J. and Shen, Zuowei},
	journal = {SIAM Journal on Optimization},
	volume = {20},
	number = {4},
	pages = {1956-1982},
	year = {2010},
	doi = {10.1137/080738970},
}

@article{dataset,
	title={Deep Learning Techniques for Automatic MRI Cardiac Multi-structures Segmentation and Diagnosis: Is the Problem Solved ?}, 
	author={O. Bernard and A. Lalande and C. Zotti, F. Cervenansky, et al.},
	year={2018},
	journal={IEEE Transactions on Medical Imaging},
	volume={37},
	pages={2514-2525}
}

@article{biobank,
	title={Imaging in population science: cardiovascular magnetic resonance in 100,000 participants of UK Biobank - rationale, challenges and approaches.}, 
	author={Petersen, S.E. and Matthews, P.M. and Bamberg, F. et al.},
	year={2013},
	journal={Cardiovascular Magnetic Resonance},
	volume={15}
}

@article{variablecontrast,
	title={Diffeomorphic registration of images with variable contrast enhancement}, 
	author={Janssens, G. and Jacques, L. and Orban de Xivry, J. and Geets, X. and Macq, B.},
	year={2011},
	journal={International journal of biomedical imaging}
}

@ARTICLE{lowranksparsecardiac,
	author={Vaswani, Namrata and Bouwmans, Thierry and Javed, Sajid and Narayanamurthy, Praneeth},
	journal={IEEE Signal Processing Magazine}, 
	title={Robust Subspace Learning: Robust PCA, Robust Subspace Tracking, and Robust Subspace Recovery}, 
	year={2018},
	volume={35},
	number={4},
	pages={32-55},
	doi={10.1109/MSP.2018.2826566}
}

@article{fechter2020shot,
	title={One Shot Learning for Deformable Medical Image Registration and Periodic Motion Tracking}, 
	author={Tobias Fechter and Dimos Baltas},
	journal={IEEE Transactions on Medical Imaging},
	year={2020},
}

@article{hering2018enhancing,
	title={Enhancing Label-Driven Deep Deformable Image Registration with Local Distance Metrics for State-of-the-Art Cardiac Motion Tracking}, 
	author={Alessa Hering and Sven Kuckertz and Stefan Heldmann and Mattias Heinrich},
	journal={Bildverarbeitung f{\"u}r die Medizin},
	pages={309-314},
	year={2018},
}

@article{pcalungmotion,
	title={On a PCA-based lung motion model.},
	author={Li R and Lewis JH and Jia X et al.},
	year={2011},
	journal={Physics in medicine and biology},
	volume={56},
	pages={6009-6030}
}

@article{cardiacstrain,
	title={Myocardial strain imaging: how useful is it in clinical decision making?},
	author={Smiseth, Otto and Torp, Hans and Opdahl, Anders and Haugaa, Kristina and Urheim, Stig},
	year={2015},
	journal={European heart journal},
	volume={37},
	pages={1196-1207}
}

@article{ejvsstrain,
	author = {Halliday, Brian P and Senior, Roxy and Pennell, Dudley J},
	title = "{Assessing left ventricular systolic function: from ejection fraction to strain analysis}",
	journal = {European Heart Journal},
	volume = {42},
	number = {7},
	pages = {789-797},
	year = {2020},
	eprint = {https://academic.oup.com/eurheartj/article-pdf/42/7/789/46625254/ehaa587.pdf},
}
\appendix
\chapter{Source Code}

A large portion of my code can be found at - https://git.cs.bham.ac.uk/sxr541/dissertation - this will not be all of my code as the datasets we made use of are very large and cannot all be uploaded. For this reason I upload all of my functions under file name "myfunctions.py". Then I uploaded my jupyter notebook "running.ipynb" which has the code already run so you can see the outputs. To run the code just open the "running.ipynb" file which contains multiple examples that we discussed throughout this paper. You can then just run each kernel independently as long as you run the imports kernel first. Or you can just look at the outputs currently saved in the notebook. For example, you can see the DiffIR and motion tracking examples on; pairwise circle to c registration; a pairwise cardiac examples; a group-wise cardiac example. \\

\textbf{IMPORTANT: I take no credit for the following functions as these were written by my supervisor, Dr. Jinming Duan:} deform\textunderscore to\textunderscore hsv, deform\textunderscore to\textunderscore disp\textunderscore 2D, central\textunderscore finite\textunderscore diff, warp\textunderscore 2D\textunderscore deformation, compose\textunderscore 2D\\

\textbf{IMPORTANT: I take no credit for the following function as this was found online and all credit goes to the original writer:} plot\textunderscore grid - This was found online at: \url{https://stackoverflow.com/questions/47295473/how-to-plot-using-matplotlib-python-colahs-deformed-grid}\\

\textbf{IMPORTANT: The following functions were written in Matlab and given to me by my supervisor Dr. Jinming Duan, but I converted them to Python so they are compatible with Pytorch:} diffIRADMM, pyramid\textunderscore flow\\

\textbf{The following functions I have written myself:} channels, Dice, padarray, zeropadarray8, rescale\textunderscore intensities, shrink, svd\textunderscore threshold, motionTrackingADMM, pyramid\textunderscore flow\textunderscore motion - the shrink and svd\textunderscore threshold functions follow the maths set out in \cite{candes2009robust} and motionTrackingADMM, pyramid\textunderscore flow\textunderscore motion are my own work, however, while the maths behind the method is my own work, it is a new method adapted from the diffeomorphic image registration algorithm set out in \cite{thorley2021nesterov} and as such the code is adapted from the functions: diffIRADMM, pyramid\textunderscore flow

\end{document}